\documentclass[lettersize,journal]{IEEEtran}
\usepackage{amsmath,amsfonts}
\usepackage{algorithmic}
\usepackage{algorithm}
\usepackage{array}
\usepackage[caption=false,font=normalsize,labelfont=sf,textfont=sf]{subfig}
\usepackage{textcomp}
\usepackage{stfloats}
\usepackage{url}
\usepackage{verbatim}
\usepackage{graphicx}
\usepackage{cite}
\usepackage{multirow}
\usepackage{booktabs}
\usepackage[normalem]{ulem}
\usepackage{pifont}
\usepackage{bbm}
\usepackage{caption}
\usepackage{afterpage}
\usepackage[marginal]{footmisc}
\usepackage{hyperref}

\begin{document}

\title{New Dataset and Methods for Fine-Grained Compositional Referring Expression Comprehension via Specialist-MLLM Collaboration}

% \author{
% Xuzheng~Yang,
% Junzhuo~Liu,
% Peng~Wang,
% Guoqing~Wang,~\IEEEmembership{Member,~IEEE,} 
% Yang~Yang,~\IEEEmembership{Senior Member,~IEEE,} 
% and~Heng~Tao~SHEN,~\IEEEmembership{Fellow,~IEEE}
\author{
	\IEEEauthorblockN{
        Xuzheng~Yang\IEEEauthorrefmark{1}\textsuperscript{1},
        Junzhuo~Liu\IEEEauthorrefmark{1}\textsuperscript{1},
        Peng~Wang\IEEEauthorrefmark{2}\textsuperscript{1},
        Guoqing~Wang\textsuperscript{1},~\IEEEmembership{Member,~IEEE,} 
        Yang~Yang\textsuperscript{1},~\IEEEmembership{Senior Member,~IEEE,} 
        and~Heng~Tao~Shen\textsuperscript{1}\textsuperscript{2},~\IEEEmembership{Fellow,~IEEE}}
        
	\IEEEauthorblockA{\textsuperscript{1}University of Electronic Science and Technology of China}
	\IEEEauthorblockA{\textsuperscript{2}Tongji University}
    \thanks{X. Yang and J. Liu contributed equally to this work. P. Wang is the corresponding author.}
} 

\markboth{Journal of \LaTeX\ Class Files,~Vol.~14, No.~8, August~2021}%
{Shell \MakeLowercase{\textit{et al.}}: A Sample Article Using IEEEtran.cls for IEEE Journals}

\IEEEpubid{0000--0000/00\$00.00~\copyright~2021 IEEE}

\maketitle
\begin{abstract}
Referring Expression Comprehension (REC) is a foundational cross-modal task that evaluates the interplay of language understanding, image comprehension, and language-to-image grounding. It serves as an essential testing ground for Multimodal Large Language Models (MLLMs). To advance this field, we introduced a new REC dataset in our previous conference paper, characterized by two key features. First, it is designed with controllable difficulty levels, requiring multi-level fine-grained reasoning across object categories, attributes, and multi-hop relationships. Second, it incorporates negative text and images generated through fine-grained editing and augmentation, explicitly testing a model’s ability to reject scenarios where the target object is absent—an often-overlooked yet critical challenge in existing datasets. In this extended work, we propose two new methods to tackle the challenges of fine-grained REC by combining the strengths of Specialist Models and MLLMs. The first method adaptively assigns simple cases to faster, lightweight models and reserves complex ones for powerful MLLMs, balancing accuracy and efficiency. The second method lets a specialist generate a set of possible object regions, and the MLLM selects the most plausible one using its reasoning ability. These collaborative strategies lead to significant improvements on our dataset and other challenging benchmarks. Our results show that combining specialized and general-purpose models offers a practical path toward solving complex real-world vision-language tasks. Our dataset and code are available at \href{https://github.com/sleepyshep/FineCops-Ref}{https://github.com/sleepyshep/FineCops-Ref}.

\end{abstract}

\begin{IEEEkeywords}
Referring Expression Comprehension (REC), Multimodal Large Language Models (MLLMs), Specialist Model, reasoning
\end{IEEEkeywords}

\section{Introduction}
\IEEEPARstart{R}{eferring} Expression Comprehension (REC) is a fundamental vision-to-language task that aims to localize a target object described by a natural language expression~\cite{xiao2024visualgroundingsurvey}. Successfully addressing this task requires visual understanding, linguistic comprehension, and language-to-image grounding, making it an ideal testbed for Multimodal Large Language Models (MLLMs). REC has broad applications in assistive AI systems, such as a robotic assistant identifying ``the red cup next to the coffee make'' in a cluttered kitchen or an AI-powered medical tool pinpointing a specific lesion in radiology scans based on textual descriptions from doctors.

To advance Referring Expression Comprehension research and its real-world applications, datasets tailored for Multimodal Large Language Models (MLLMs) are essential. Traditional REC benchmarks such as RefCOCO, RefCOCO+, and RefCOCOg~\cite{mao2016generation, yu2016modeling, Nagaraja2016modeling} lack considerations for compositional reasoning, allowing models to perform well without truly understanding linguistic structure—or in some cases, even without relying on the expression itself~\cite{cirik-etal-2018-visual,akula-etal-2020-words}. Moreover, most existing datasets do not include negative samples, where the target object is absent from the image. This omission is critical, as real-world REC scenarios often require models to reject incorrect object proposals when the described entity is not present. The lack of such negative examples limits the robustness of current REC models, making them prone to false positives.
\IEEEpubidadjcol

To address these limitations in existing datasets, we introduced the FineCops-Ref benchmark in our previous work~\cite{liu2024finecops}. FineCops-Ref is specifically designed to enhance fine-grained reasoning in REC by incorporating controlled difficulty levels and meticulously constructed negative samples.
The difficulty levels compel MLLMs to reason across object categories, attributes, and multi-hop relationships, categorized based on the number of attributes and relationships necessary for accurate localization. The inclusion of negative samples evaluates model resilience against misalignments and hallucinations, providing a more rigorous assessment of their true visual grounding capabilities.
By addressing both compositional reasoning and negative sample inclusion, FineCops-Ref establishes a more realistic and challenging benchmark for advancing REC research in the MLLM era.

Extensive evaluations on the FineCops-Ref benchmark also reveal a critical gap in current vision-language models~\cite{liu2024finecops}: while MLLMs demonstrate strong generalization and reasoning abilities, they often struggle with fine-grained localization and compositional grounding, particularly when precise object differentiation is required. In contrast, Specialist Models, which are typically lightweight and trained for specific vision tasks, excel in efficient and accurate low-level perception but lack the broad contextual reasoning capabilities of MLLMs. Given their complementary strengths, an effective solution is to leverage both models strategically—assigning simpler, perception-driven tasks to Specialist Models while reserving MLLMs for complex reasoning. This collaborative approach not only enhances accuracy and robustness but may also improve computational efficiency, as MLLMs are invoked only when necessary. 

Building on this insight, this extended work introduces two collaboration strategies between Specialist Models and MLLMs to improve REC performance. The first strategy, Slow-Fast Adaptation (SFA), introduces an adaptive task routing mechanism that dynamically assigns simpler, detection-oriented tasks to lightweight Specialist Models, while delegating complex tasks requiring compositional reasoning to MLLMs. This adaptive and flexible collaboration not only improves efficiency but, more importantly, enhances overall REC performance. Additionally, we identify common error patterns in both types of models, particularly their tendency to mistakenly identify objects not belonging to the described category. To mitigate this issue, we propose a simple yet effective target-refocus strategy, which obviously reduces localization confusion for both Specialist Models and MLLMs. Notably, the SFA strategy is entirely training-free, making it highly practical for real-world applications. The second strategy, Candidate Region Selection (CRS), aims to improve recall and precision by generating multiple object candidates using Specialist Models, followed by leveraging MLLMs' reasoning capabilities to select the correct target. While MLLMs inherently possess reasoning capacity, they are not explicitly trained to approach selection tasks as generative problems, limiting their effectiveness in such scenarios. To address this, we conduct instruction tuning on external datasets, ensuring that MLLMs can effectively address selection tasks while avoiding knowledge leakage. This instruction tuning process generalizes well across various REC datasets.

Experiments on three compositional REC datasets demonstrate that our Specialist-MLLM collaboration strategies significantly enhance the performance of both existing Specialist Models and MLLMs, highlighting the generalizability of our approach. Beyond achieving superior performance, we aim for this work to provide valuable insights into solving complex real-world tasks by strategically leveraging existing models for maximum effectiveness, rather than reinventing them. This aligns with the broader trend of rapidly evolving LLM-based agents, emphasizing the importance of optimizing their potential to tackle challenging tasks efficiently.

This work extends our previous conference paper~\cite{liu2024finecops}, which introduced the FineCops-Ref benchmark. In this journal version, we make three major contributions: (1) we propose two novel Specialist–MLLM collaboration strategies tailored for fine-grained compositional REC; (2) we conduct more extensive experiments across FineCops-Ref and multiple external benchmarks; and (3) we provide deeper analyses and ablations, offering a more self-contained and comprehensive exploration of the task.

%Motivated by this insight, we propose a Specialist-MLLM collaboration strategy, which introduces adaptive task routing and candidate selection mechanisms to optimize performance, efficiency, and generalization in REC.

%From a methodological perspective, MLLMs have demonstrated strong linguistic and knowledge capabilities, achieving impressive results across various vision-to-language tasks. However, they struggle with fine-grained compositional cross-modal reasoning, which is essential for understanding complex referring expressions. For instance, MLLMs may misinterpret hierarchical relationships in descriptions such as ``the book on the left side of the shelf above the desk'', failing to correctly resolve multi-step spatial dependencies. Additionally, their tendency to rely on high-level correlations rather than precise object attributes makes them less effective in distinguishing between similar objects with subtle differences. Furthermore, the high computational cost of MLLMs poses a significant challenge, limiting their practical deployment in real-world applications.

\section{Related Works}
\subsection{Benchmarking Referring Expression Comprehension}
Referring Expression Comprehension (REC) is a vision-language task where models must localize a target object in an image based on a natural language referring expression, requiring compositional understanding of attributes (e.g., color, size), spatial relationships, and object categories to disambiguate the referent from other entities. While early benchmarks like RefCOCO/+/g~\cite{mao2016generation,yu2016modeling,Nagaraja2016modeling} established foundational evaluation protocols, subsequent analyses revealed critical limitations. For instance, \cite{cirik-etal-2018-visual} and \cite{akula-etal-2020-words} demonstrated that models could exploit dataset biases rather than genuinely parse linguistic structure: \cite{akula-etal-2020-words} found that up to 83.7\% of RefCOCOg test instances could be solved via keyword matching or visual feature dominance, bypassing compositional reasoning. To address this, \cite{akula-etal-2020-words} introduced Ref-Adv, a dataset where perturbed expressions refer to alternate objects, forcing models to engage with linguistic structure. Subsequent efforts have prioritized compositional reasoning. CLEVR-Ref+~\cite{liu2019clevr} is a synthetic dataset emphasizing relationships, attributes, and linguistic logic. 
Cops-Ref~\cite{chen2020cops} and Ref-Reasoning~\cite{yang2020graph} use GQA scene graphs~\cite{hudson2019gqa} and rule-based methods to create large-scale compositional referring expression comprehension datasets in real-world scenarios. Cops-Ref additionally introduces distracting images based on attributes, relationships, and target names. Recent benchmarks like GITM-MR~\cite{Wu_2023_ICCV_GITM-MR} explores mismatched relationship in the REC task, while ARPGrounding~\cite{zeng2024investigating} evaluates the model's compositional reasoning ability by constructing referring expression pairs that target distinct objects of the same category within a single image, differentiated solely through their attributes or relational contexts. RefEgo~\cite{kurita2023refego} and OmniLabel~\cite{schulter2023omnilabel} consider out-of-distribution scenarios where the referred targets do not exist in the image.

The challenges in rigorously evaluating compositional reasoning are not unique to REC. Across various related vision-language tasks, such as image captioning, image-to-text retrieval, and visual question answering, recent evaluations similarly reveal that current vision-Language Models struggle with compositional understanding~\cite{shekhar-etal-2017-foil,parcalabescu-etal-2022-valse,thrush2022winoground,ma2023crepe,NEURIPS2023_coco_counter,Kamath2024TheHP,hsieh_sugarcrepe_2023,yarom_what_2023,tong2024eyes,zhang2024countercurate,lin2023visualgptscore,coco_cf}. For instance, prominent benchmarks like FOIL-it~\cite{shekhar-etal-2017-foil}, VALSE~\cite{parcalabescu-etal-2022-valse}, and COCO-Counterfactuals~\cite{NEURIPS2023_coco_counter} are specifically designed to test compositional understanding by creating ``hard negative'' image-caption pairs that highlight subtle distinctions. These broader evaluation efforts underscore the critical need for benchmarks that can truly test deep linguistic-visual understanding and resilience to subtle distractors in complex, compositional scenarios – a requirement particularly pronounced in fine-grained REC where minor linguistic cues can drastically change the target.

Building on these advances, we introduced FineCops-Ref in our previous work~\cite{liu2024finecops}, a new REC benchmark specifically designed to more comprehensively assess the fine-grained compositional reasoning abilities of multimodal models.
Based on this challenging benchmark, we proposes novel Specialist-MLLM collaboration strategies and provides more extensive experiments to further advance fine-grained compositional REC.

% Despite advancements in multimodal learning,
% current multimodal models, including advanced MLLMs like GPT-4V, exhibit poor compositional reasoning, often treating language as a bag of words without considering word order, attributes, or relationships between objects~\cite{suhr-etal-2019-corpus,ma2023crepe,diwan-etal-2022-winoground,tong2024eyes,yuksekgonul2022and,coco_cf,Kamath2024TheHP}. To evaluate these models of compositional reasoning, benchmarks often involve constructing hard negative captions to test models' capabilities, such as distinguishing between "a mug in some grass" and "some grass in a mug"~\cite{parcalabescu-etal-2022-valse,thrush2022winoground,ma2023crepe,hsieh_sugarcrepe_2023}. 
% Some benchmarks focus on negative images~\cite{ray_cola_2023,yarom_what_2023,zhang-etal-2024-countercurate,NEURIPS2023_coco_counter}, while others primarily focus on spatial relationships~\cite{zhang-etal-2024-countercurate,liu-etal-2023-visual,yang2019spatialsense,chen2024spatialvlm}.
% %\cite{hsieh_sugarcrepe_2023} found that previous benchmarks have language biases and that a simple grammar model can distinguish negative captions. 

\subsection{Specialist REC Models}
% The REC methods can generally be divided into two categories based on whether or not it uses LLMs: specialist models and MLLMs.
Early Specialist Models~\cite{hu2017modeling, zhang2018grounding, zhuang2018parallel, yu2018mattnet, liu2019improving, yang2019dynamic, hong2019learning, liu2019learning,Wang2018NeighbourhoodWR} typically employed pre-trained detectors to generate proposals and locate targets through text-to-region matching. Subsequent approaches transitioned to ViT-based architectures, typically adopting one-stage designs~\cite{zhou2021real, deng2021transvg, zhu2022seqtr, TransVG++2023}.
Recent advancements~\cite{kamath2021mdetr, yan2023universal, liu2023grounding, dMDETR23} have shifted towards DETR-based~\cite{carion2020end} frameworks, utilizing encoder-decoder architectures for multimodal fusion and generating target bounding boxes through object queries. Specifically, MDETR~\cite{kamath2021mdetr} pioneered the integration of free-form text with DETR, developing an end-to-end modulated detector capable of identifying objects in images based on raw text queries. UNINEXT~\cite{yan2023universal} implemented Deformable DETR~\cite{zhu2020deformable}, proposing a universal instance perception model. Grounding-DINO~\cite{liu2023grounding}, building upon the advanced DINO framework~\cite{zhang2022dino}, enhanced DINO by incorporating vision-language modality fusion at multiple stages. \cite{Zhao_2024_CVPR} explored the utilization of hard negative samples during training.
More recent models, such as OneRef~\cite{xiao2024oneref} and SimVG~\cite{dai2024simvg} have streamlined the architecture by eliminating complex integration designs and redundant parameters through the implementation of modality-shared feature spaces, employing a unified multimodal encoder. Specialist REC models, typically lightweight and specifically trained for vision tasks, excel in efficient and accurate low-level perception. However, they generally lack broad contextual reasoning capabilities, which limits their performance on complex REC tasks.

\begin{figure*}[!t]
  \centering
  \includegraphics[width=\textwidth]{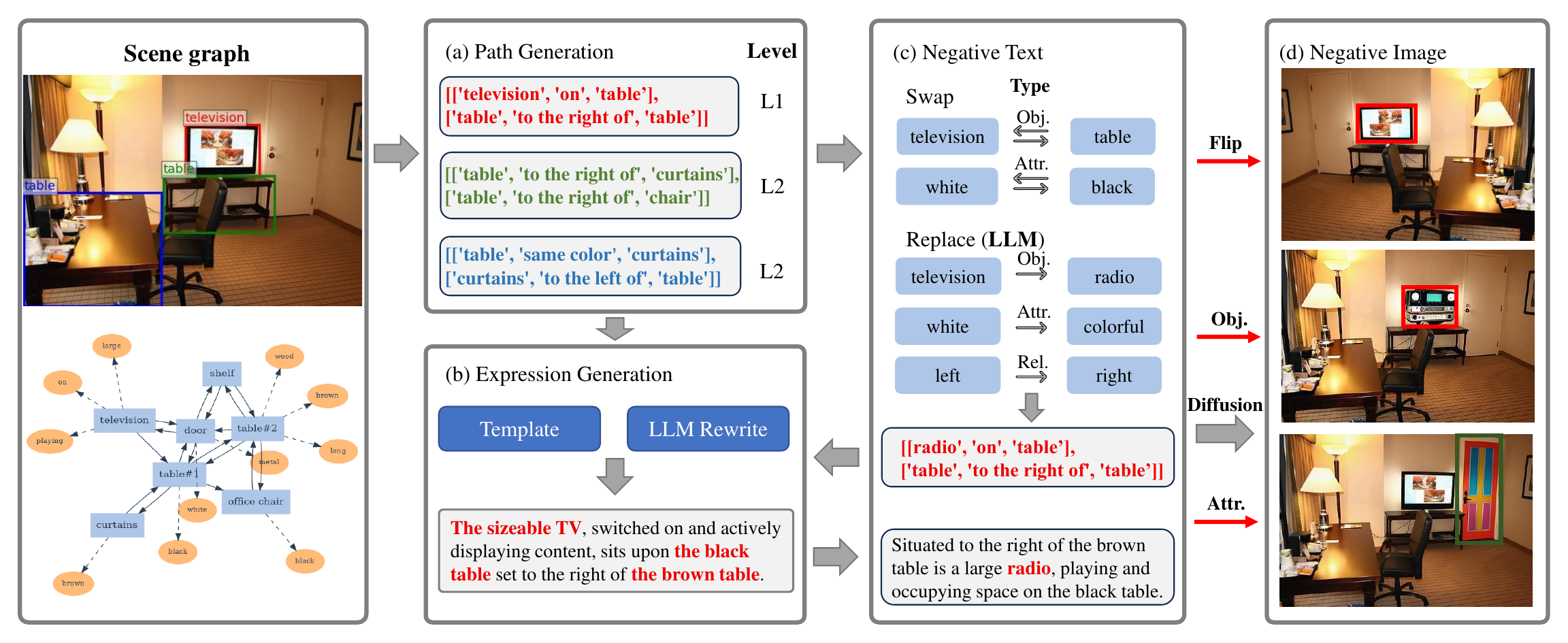}
  \caption{The data construction pipeline of FineCops-Ref~\cite{liu2024finecops}. Given an image, we first generate paths based on its scene graph. Then, we fill paths into templates and obtain the positive referring expression through LLM rewriting. Meanwhile, we utilize LLM to generate negative expressions, and based on this, we employ diffusion model to create fine-grained editing negative images.}
  \label{pipeline}
\end{figure*}

\subsection{Multimodal Large Language Models for REC}
Unlike Specialist REC models, which usually treat REC as a bounding box regression task, MLLMs often formulate the bounding box prediction as a text generation problem, outputing bounding box coordinates in an autoregressive manner.
Recent MLLMs~\cite{chen2023shikra, you2024ferret, li-etal-2024-groundinggpt, wei2023lenna, wang2023cogvlm, qi2024cogcom, chen2024far, yang2023set} have leveraged the powerful capabilities of LLMs through zero-shot or instruction tuning methods to address complex REC tasks. These MLLMs also known as grounding MLLMs~\cite{xiao2024visualgroundingsurvey}.
%These methods integrate projected visual features with text tokens into LLM, which then directly outputs the bounding box.
Shikra~\cite{chen2023shikra} is an early model to support region-level input and understanding. 
% Ferret~\cite{you2024ferret} offers more flexible referring by implementing a hybrid representation scheme. 
GroundingGPT~\cite{li-etal-2024-groundinggpt} employs a coarse-to-fine training strategy to enhance model’s semantic awareness. CogVLM~\cite{wang2023cogvlm} incorporated a visual expert module in each transformer layer to enable dual fusion between vision and language features. CogCoM~\cite{qi2024cogcom} introduced Chain of Manipulations, encouraging MLLMs to generate responses through evidential visual reasoning. Regarding visual modules, some methods, such as GLaMM~\cite{Rasheed_2024_CVPR} and LLaVA-Grounding~\cite{zhang2023llava-g}, integrate additional visual components, while others, including Groma~\cite{ma2024groma}, VisualCoT~\cite{shao2024visualcot}, Ferret~\cite{you2024ferret}, and CoVLM~\cite{li2024covlm}, extract regional features as supplementary inputs. Recently, generalist MLLMs, such as Qwen2-VL~\cite{wang2024qwen2} and InternVL 2.5~\cite{chen2024expanding}, have demonstrated significant advancements in grounding capabilities while maintaining strong reasoning performance, achieved through enhanced model architectures, carefully curated datasets, and optimized training strategies.

% However, MLLMs employ LLMs with a vast number of parameters and FLOPs, often exceeding specialist models by tens to hundreds of times, resulting in substantial computational costs. Additionally, MLLMs project visual tokens into the input space of LLMs, increasing the input length. This results in computational complexity growing quadratically with the combined number of textual and visual tokens, also significantly amplifying the computational burden. Despite these computational limitations, MLLMs offer superior generalization and reasoning abilities compared to specialist models, making them suitable for complex REC tasks and scenarios requiring broader contextual understanding.

MLLMs demonstrate superior generalization and reasoning capabilities compared to Specialist Models, making them particularly well-suited for complex REC tasks demanding deeper contextual understanding. However, MLLMs inherit significant computational burdens from two key aspects. First, they rely on LLMs which possess substantially more parameters than Specialist Models. Second, the attention mechanisms used by MLLMs exhibit quadratic scaling of computational complexity. This scaling is further exacerbated by the increased input length resulting from the integration of numerous visual tokens (e.g., image patches) with textual tokens.

% MLLMs directly input the projected visual features into the LLM.
% Recent methods aim to enhance grounding capabilities in MLLMs through dataset construction with coordinate information and additional visual modules. Common methods for datasets include transforming traditional visual datasets into an instruction-following format using templates~\cite{li-etal-2024-groundinggpt,pramanick2023jack,wang_visionllm_2023}, correlating object coordinates with existing captions~\cite{peng2024grounding,qi2024cogcom}, and using LLMs to generate grounded question-answer pairs based on images, object coordinates, and captions~\cite{you2024ferret, wang2024the}. 
% % The All-Seeing Project~\cite{wang2024the} has recently introduced a new dataset (AS-1B) using a scalable data engine that incorporates human feedback and efficient models in the loop.

% In terms of visual modules, some methods integrate additional visual components, such as GLaMM~\cite{Rasheed_2024_CVPR} and LLaVA-Grounding~\cite{zhang2023llava-g}, while others extract regional features as additional inputs~\cite{ma2024groma,shao2024visualcot,you2024ferret,li2024covlm}.

\section{FineCops-Ref Benchmark}
In this section, we provide a summary of the FineCops-Ref dataset introduced in our previous work~\cite{liu2024finecops}. FineCops-Ref is a benchmark specifically designed to address the limitations of existing datasets in evaluating fine-grained compositional reasoning and robustness to negative samples in Referring Expression Comprehension. It comprises both positive and negative samples, carefully constructed to provide a more challenging and realistic evaluation setting for Multimodal models. The dataset construction pipeline is illustrated in Fig. 1. The detailed methodology for constructing FineCops-Ref, including path generation, data categorization, expression generation, human filtering, and the generation of negative data (both expressions and images), was introduced in our previous conference work~\cite{liu2024finecops}.

\subsection{Key Characteristics}
FineCops-Ref is distinguished by the following key characteristics, which make it a suitable and challenging benchmark for evaluating advanced REC models and our proposed collaboration strategies:

\textbf{Controlled Difficulty Levels:} The dataset incorporates three controlled difficulty levels (Level 1, 2, and 3) based on the complexity of fine-grained reasoning required to uniquely identify the target object. This necessitates models to reason across object categories, attributes, and multi-hop relationships. Level 1 involves simple localization without distractors of the same category, Level 2 requires distinguishing the target using a single attribute or relation among same-category distractors, and Level 3 demands the interpretation of two or more relationships and attribute dependencies.

\textbf{Comprehensive Negative Samples:} FineCops-Ref includes both negative expressions and negative images. These negative samples are specifically designed as ``hard negatives'' with subtle differences from positive instances, evaluating the model's ability to correctly reject instances where the described object is absent or the expression/image is subtly mismatched. The construction involves various types of challenging negative samples generated through careful editing and perturbation.

% FineCops-ref comprises both positive and negative data, designed to evaluate fine-grained visual reasoning. This section details the process for creating the positive and negative data.
% FineCops-Ref includes both positive and negative data. Fig.~\ref{pipeline} illustrates the dataset construction pipeline.

\begin{table*}[t]
\caption{\label{benchmark_comparison}
Comparison between the proposed benchmark and other REC benchmarks. Unconstrained  indicates the final expression is not constrained by the templates. Cops. indicates compositional reasoning. On the right hand side, the test set count of each benchmark is listed.
}
\centering
\resizebox{\textwidth}{!}{
\begin{tabular}{lcccccccc} 
\toprule
                   &                   &                           &                           &                   &                    & {Positive}   & \multicolumn{2}{c}{{Negative}}  \\ 
\cmidrule(lr){7-7}
\cmidrule(lr){8-9}
{Benchmark} & {Unconstrained} & {Cops.} & {Difficulty level} & {Neg. text} & {Neg. image} & {Expression} & {Expression} & {Image}    \\ 
\midrule
RefCOCO~\cite{yu2016modeling}            & \ding{51}                 &                           &                           &                   &                    & 10,752               & -                   & -                 \\
RefCOCO+~\cite{yu2016modeling}            & \ding{51}                 &                           &                           &                   &                    & 10,615               & -                   & -                 \\
RefCOCOg~\cite{Nagaraja2016modeling}           & \ding{51}                 &                           &                           &                   &                    & 9,602               & -                   & -                 \\
Ref-Reasoning~\cite{yang2020graph}      &                   & \ding{51}                         & \ding{51}                         &                   &                    & 34,609              & -                   & -                 \\
Cops-ref~\cite{chen2020cops}           &                   & \ding{51}                         &                           &                   & \ding{51}                  & 12,586               & -                   & 37,758             \\
Ref-Adv~\cite{akula-etal-2020-words}            & \ding{51}                 & \ding{51}                         & \ding{51}                         & \ding{51}                 &                    & 9,602                & 3,704                & -                 \\
Ours                & \ding{51}                 & \ding{51}                         & \ding{51}                         & \ding{51}                 & \ding{51}                  & 9,605                & 9,814                & 8,507              \\
\bottomrule
\end{tabular}}
\end{table*}

\begin{table}
\caption{\label{dataset_statics_all}
Dataset statistics.}
\centering
\resizebox{\columnwidth}{!}{
\begin{tabular}{cccc} 
\toprule
{Set}   & {Positive} & {Negative expression} & {Negative image}  \\ 
\midrule
Train & 163,792   & 80,451               & -               \\
Val   & 18,455    & 9,029                & -               \\
Test  & 9,605     & 9,814                & 8,507            \\
\bottomrule
\end{tabular}}
\end{table}

\subsection{Dataset Statistics}
FineCops-Ref consists of 9,605 positive expressions, 9,814 negative expressions, and 8,507 negative images in test set. Table~\ref{benchmark_comparison} provides a comparison between FineCops-Ref and other visual grounding benchmarks. FineCops-Ref combines the advantages of unconstrained expression, fine-grained compositional reasoning, difficulty level, and hard negatives at both textual and visual levels. Additionally, we partition the training set and validation set simultaneously as in Table~\ref{dataset_statics_all}. For more statistics, please refer to the Appendix.

\subsection{Metrics}

To evaluate performance on positive data, we use the common metric Precision@k.  
When both positive and negative data are present in the test set, we treat the negative samples as distractors for the positive samples, and introduce two additional metrics:

{Recall@k}:  
We treat the REC task as a bounding box retrieval problem. For each negative sample paired with its corresponding positive sample, we first obtain the predicted bounding boxes from the model, along with their confidence scores for both positive and negative samples. These bounding boxes are then ranked based on their confidence scores.  
Recall@k measures the proportion of negative-positive pairs where at least one of the top \(k\) predicted bounding boxes has an IoU greater than 0.5 with the ground truth bounding box. It specifically assesses the model's ability to avoid assigning high confidence scores to negative samples. Formally, Recall@k is defined as:

{\small
\begin{equation}
\text{Recall@k} = \frac{1}{N} \sum_{i=1}^{N} \mathbbm{1}\left( \max_{j \in \{1, \ldots, k\}} \text{IoU}_{i,j} > 0.5 \right),
\end{equation}
}
where \( N \) represents the total number of negative-positive pairs, and \( \mathbbm{1}(\cdot) \) is an indicator function that equals 1 if the condition inside is true and 0 otherwise. The term \(\text{IoU}_{i,j}\) refers to the overlap between the \(j\)-th predicted bounding box (ranked based on the confidence scores) and the ground truth bounding box for the \(i\)-th pair. Note that for negative samples, there is no ground truth box, meaning the IoU is 0.

Recall@k is commonly used in retrieval tasks to assess prediction accuracy in the presence of challenging negative samples. Ideally, the model should assign lower confidence scores to negative samples. In our study, we primarily report Recall@1. If the model consistently assigns lower confidence scores to negative samples compared to positive ones, Recall@1 should equal Precision@1.

{AUROC}:  
While Recall@k evaluates how well the model ranks individual negative samples relative to their corresponding positive samples, it does not offer a holistic view of confidence across the dataset. To address this, we use AUROC to measure the overall ability of the model to distinguish between positive and negative samples.  
AUROC measures the model's ability to correctly rank positive samples higher than negative ones across the datasets, providing a holistic view of its discriminative power.

By combining Recall@k and AUROC, we ensure a comprehensive evaluation of the model's performance in distinguishing between positive and negative samples in REC tasks. This dual approach addresses both specific ranking and overall confidence.

\begin{figure*}
    \centering
    \includegraphics[width=0.9\textwidth]{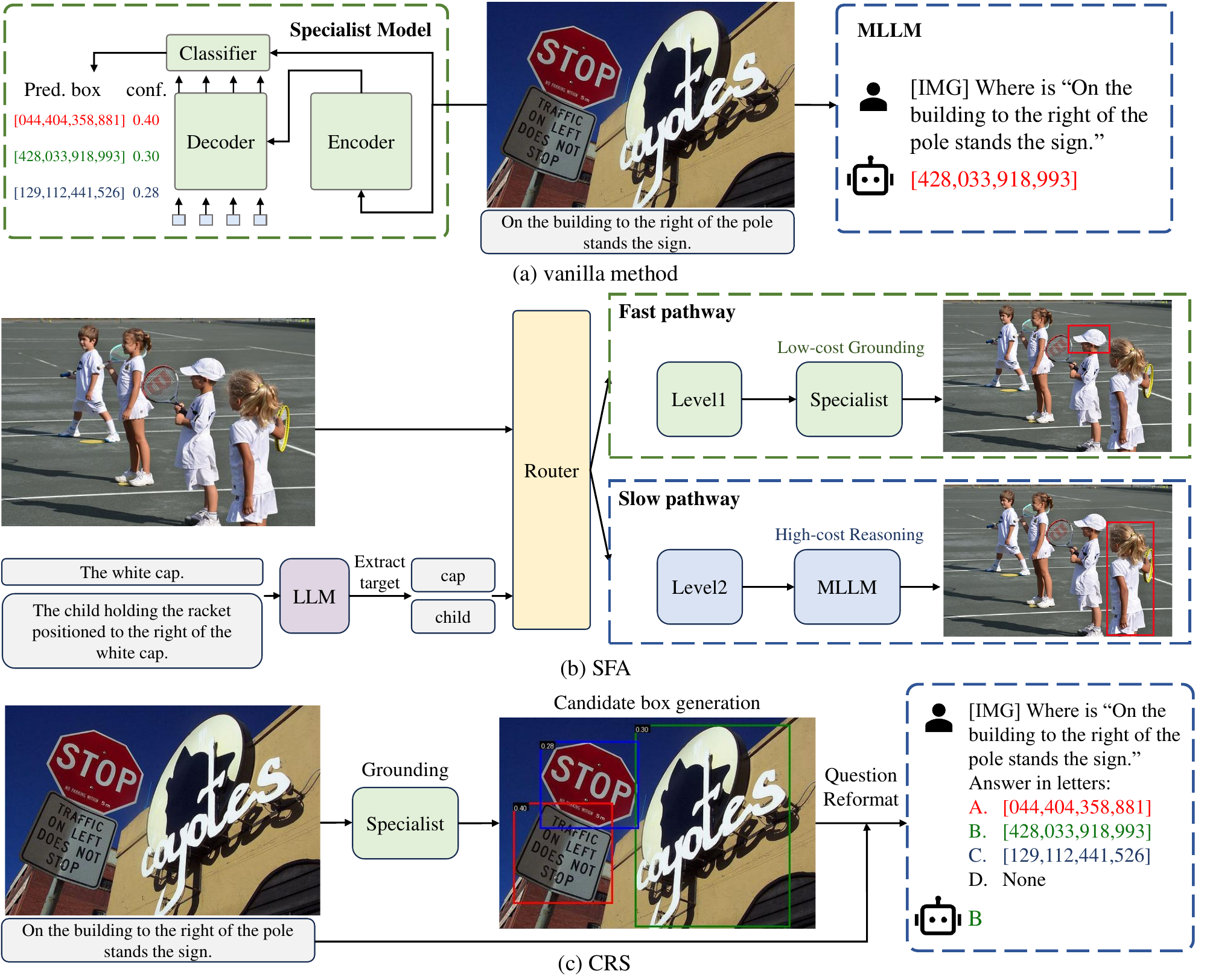}
    \caption{(a) Vanilla Model: The Specialist Model uses box regression to output multiple candidate boxes with confidence scores and selects the region with the highest confidence. In contrast, MLLMs directly generate object coordinates as text.
(b) Slow-Fast Adaptation (SFA): A router categorizes tasks by complexity (Level 1/2). Level 1 tasks are processed using the fast pathway (Specialist Model), while Level 2 tasks are processed using the slow pathway (MLLM).
(c) Candidate Region Selection (CRS): The Specialist Model generates multiple candidate bounding boxes, and the MLLM selects the best option in a multi-choice selection task.}
    \label{fig_method}
\end{figure*}

\section{Method}
Referring Expression Comprehension presents a unique challenge in computer vision, demanding not only precise object detection but also compositional visual reasoning to ground natural language descriptions to specific image regions. Specialist Models, such as MDETR~\cite{kamath2021mdetr} and Grounding DINO~\cite{zhao2024open}, demonstrate efficiency and excel at accurate object detection and grounding. Their strengths lie in their optimized architectures and training on extensive datasets with bounding box annotations, enabling precise box regression and rapid processing. These models deliver exceptional grounding accuracy and high Recall@k, ensuring that the correct object is often among the top predictions. Indeed, for simpler REC instances where target identification relies primarily on object category, Specialist Models often suffice. However, Specialist Models exhibit limitations in handling the inherent complexity of natural language. They often struggle with referring expressions that require compositional reasoning, disambiguation among multiple objects of the same category.  

In contrast, MLLMs exhibit strong visual reasoning capabilities, making them well-suited for complex REC tasks that require deeper semantic understanding. However, they are computationally demanding and not inherently optimized for grounding tasks, as they generate bounding box coordinates directly as part of a text generation process, rather than leveraging intermediate bounding box proposals as informative priors, as specialist models do.
%And often perform poorly on object detection benchmarks~\cite{wu2024dettoolchain}.
% such as COCO

This motivates our collaborative approach, which aims to synergistically combine the strengths of Specialist Models and MLLMs while mitigating their respective weaknesses. We recognize the complementary nature of these architectures: Specialist Models offer efficiency and strong grounding capabilities, particularly valuable for precise object localization and candidate generation, while MLLMs provide the crucial reasoning abilities needed for disambiguation and complex expression understanding. By strategically integrating these model types, we can achieve superior REC performance with improved computational efficiency compared to relying solely on resource-intensive MLLMs.

Motivated by this insight, we introduce two novel strategies: \textbf{Slow-Fast Adaptation (SFA)} and \textbf{Candidate Region Selection (CRS)}.  \textbf{SFA} routes REC tasks based on their inherent complexity. Simpler tasks, where Specialist Models are adept, are processed efficiently, while more complex tasks requiring reasoning are directed to MLLMs. This dynamic routing optimizes overall computational cost without sacrificing performance. \textbf{CRS} leverages the strong object detection of Specialist Models to generate a set of candidate regions, and then employs an MLLM to reason over these candidates and select the most appropriate one. The subsequent subsections detail each method, elucidating their specific mechanisms and implementation.

%This strategy harnesses the Specialist Model's grounding prowess for proposal generation and the MLLM's reasoning for accurate disambiguation and final selection. Both strategies are designed to enhance REC performance while improving computational efficiency.  The subsequent subsections detail each method, elucidating their specific mechanisms and implementation.

\begin{figure*}[ht]
  \centering
   \includegraphics[width=0.85\linewidth]{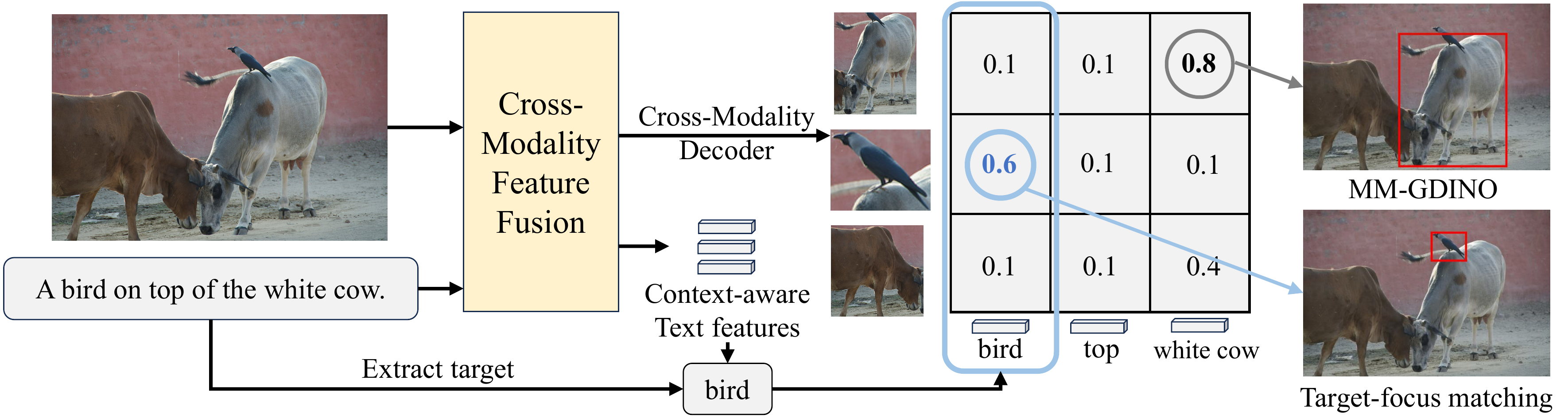}
   \caption{Illustration of the target-focus matching for Specialist Models. The target is extracted from the referring expression using a LLM. Leveraging the cross-attention mechanism, the target’s feature becomes context-aware, enabling more accurate matching with the proposal. This approach helps avoid mismatches with other objects mentioned in the expression, an issue observed in the original solution (e.g., MM-GDINO~\cite{zhao2024open}).    
   %After generating proposals, MM-GDINO~\cite{zhao2024open} typically selects the proposal with the highest similarity score with the whole expression as the output. Differently, our method focuses solely on similarity scores associated with \textless subject\textgreater \ (context-aware text feature) and selects the proposal with the highest subject score as the output. \textless subject\textgreater \ donates the subject extracted in SFA.
   }
   \label{focus_specialist}
\end{figure*}

\subsection{Slow-Fast Adaptation (SFA)}
The Slow-Fast Adaptation (SFA) strategy is designed to optimize both performance and computational efficiency in a training-free manner. As illustrated in Fig.~\ref{fig_method}(b), SFA operates by utilizing a router to evaluate task complexity and directing tasks to either a fast or slow pathway based on their difficulty level. Given an image and a referring expression, the process begins with target extraction and difficulty level assessment, followed by task routing to the appropriate pathway.

 \textbf{Target Extraction.} 
 Given a referring expression, we employ GPT-3.5-turbo to extract the target object by prompting: ``Which object does the given expression refer to?''. To minimize hallucinations by GPT, we include three in-context examples in the prompt and require GPT to respond in dictionary format. Detailed prompts are provided in the supplementary material.

 \textbf{Difficulty Level Assessment.} 
 We categorize REC tasks into two levels based on the presence of distracting objects in the image:

\begin{itemize}
    \item {Level 1:} The image contains single instance of the referred object, which can be easily identified based on the category name ( e.g., a single white cap in the image, as shown in Fig.~\ref{fig_method}(b)).
    \item {Level 2:} The image contains multiple instances of the target category, requiring compositional reasoning to disambiguate the correct target (e.g., identifying a child ``positioned to the right of the white cap'' among several children, as shown in Fig.~\ref{fig_method}(b)).
\end{itemize}

To assess task complexity, we employ a low-cost routing method: object detection is performed on the extracted target category using MM-GDINO-L~\cite{zhao2024open} with a confidence threshold of 0.2. The number of detected objects above this threshold determines the task's difficulty level. If only one object is detected, the task is classified as Level 1; otherwise, it is classified as Level 2. This approach ensures efficient routing while minimizing computational overhead.

 \textbf{Task Routing.} 
 Based on the difficulty level, tasks are routed to one of two pathways: Simple tasks are processed using low-cost Specialist Models (e.g., Grounding DINO), which make quick judgments based on the target object's category. On the other hand, complex tasks are processed using resource-intensive MLLMs, which leverage their superior reasoning capabilities to disambiguate the target object through compositional reasoning.

Notably, the SFA strategy does not require additional training and introduces minimal computational overhead. By dynamically routing tasks based on their complexity, SFA effectively combines the efficiency of Specialist Models with the reasoning ability of MLLMs, achieving higher performance while maintaining computational efficiency.

\textbf{Focus-enhancement Strategy.}
In our work, we observe that models often struggle to correctly identify the target object category referred to in an expression, frequently detecting instances from other categories mentioned in the description instead. To mitigate this issue, we propose a focus-enhancement strategy—a simple yet effective method that encourages models to precisely concentrate on the intended target category derived from the referring expression.

\textbf{Target-focus Matching for Specialist Models.} 
Fig.~\ref{focus_specialist} illustrates the focus-enhancement strategy for Specialist Models, specifically the target-focus matching mechanism. By default, MM-GDINO~\cite{zhao2024open} computes similarity scores between image tokens and all textual tokens, selecting the proposal with the highest overall score as the output. In contrast, our method refines this process by focusing solely on the similarity scores associated with the target extracted from the referring expression, ensuring that the proposal with the highest target-specific score is selected. This approach proves effective, as cross-modal feature fusion enables the target-specific token to capture both the full semantic context of the expression and the corresponding visual context. By leveraging context-aware, target-specific textual features for matching, our method filters out distracting information while preserving the essential meaning of the expression, thereby improving accuracy. As shown in Fig.~\ref{focus_specialist}, by restricting selection to scores associated with the term ``bird,'' our method avoids incorrect matches (e.g., where the score for ``white cow'' is higher) and accurately identifies the intended target.

% MM-GDINO~\cite{zhao2024open} typically selects the region proposal with the highest global score as the output. Differently, our method focuses solely on scores associated with subject and selects the proposal with the highest subject score as the output.
% Specifically, we achieve this by setting the input parameter "tokens\_positive"—used in the phrase grounding task—to the subject's position within the expression.

\begin{figure}[h]
  \centering
   \includegraphics[width=1\linewidth]{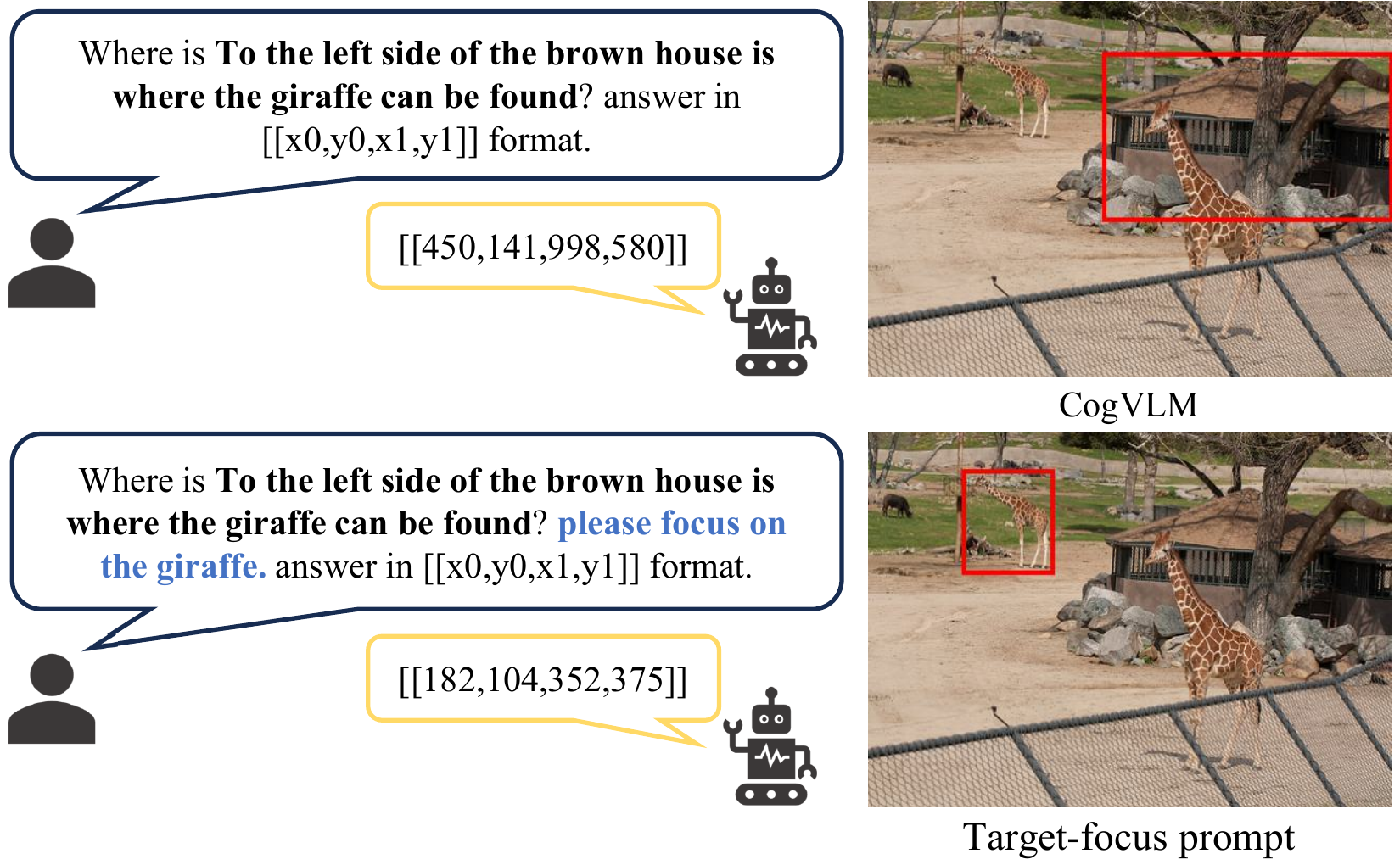}
   \caption{Target-focus prompt for MLLMs. The default prompt used by CogVLM~\cite{wang2023cogvlm} is ``Where is \textless expr\textgreater? answer in [[x0, y0, x1, y1]] format.'' as shown above. To enhance focus, we modify the prompt by appending ``please focus on the \textless target\textgreater'', where \textless target\textgreater \ donates the target extracted by LLM from the textual expression.}
   \label{focus_mllm}
\end{figure}

\textbf{Target-focus Prompt for MLLMs.} 
Fig.~\ref{focus_mllm} illustrates the focus-enhancement strategy for MLLMs, specifically the target-focus prompt. We add a simple prompt like ``please focus on the \textless target\textgreater'', where \textless target\textgreater \ denotes the extracted target from the expression. As shown in Fig.~\ref{focus_mllm}, without the target-focus prompt, CogVLM~\cite{wang2023cogvlm} incorrectly identifies the target as ``house''. However, after incorporating the target-focus prompt, the model correctly understands the intended reference ``giraffe'' and accurately locates its instance on the left side of the image. The purpose of this enhanced prompt is to explicitly emphasize the target of interest within the expression, guiding the model toward more precise interpretation and localization.

%This method enhances MLLMs' ability to develop a deeper understanding and focus on the referred target of the expression.

%Importantly, both target-focus matching and target-focus prompt are computationally inexpensive. The focus-enhancement strategy does not incurs ignorable computational costs and is entirely training-free. It effectively guides models to concentrate on the correct target, reducing the risk of misidentifying other objects.

\subsection{Candidate Region Selection (CRS)}

The Candidate Region Selection (CRS) strategy combines the complementary strengths of Specialist Models and MLLMs in a cost-effective manner. In standard REC tasks, Specialist Models output bounding boxes and typically select the highest-confidence one as the target. However, this approach can fail when the correct target is not the highest-confidence box or when multiple objects from the same category are present. By exploring the top-k bounding boxes, CRS expands the candidate set, improving the likelihood of selecting the correct target in ambiguous cases. Integrating an MLLM allows for effective disambiguation, enabling the model to reason over the top-k candidates and select the correct one through a multiple-choice format.

 \textbf{CRS Pipeline.} 
 As shown in Fig.~\ref{fig_method}(c), the CRS strategy begins by applying a Specialist Model (e.g., Grounding DINO) for object detection, generating a set of candidate bounding boxes. Non-Maximum Suppression (NMS) with a 0.7 threshold is then applied to remove redundant predictions. From the remaining boxes, the top-k highest-confidence ones are selected as potential targets.
                                                                                                                                                                                      
These candidates are passed to the MLLM, which processes them in a multiple-choice format. The MLLM selects the correct bounding box by outputting a single token (e.g., ``A''), simplifying the prediction process compared to traditional autoregressive methods that predict bounding box coordinates sequentially. This approach not only reduces inference time but also provides the MLLM with more prior information for accurate target detection.
%allowing it to focus on reasoning over a limited set of candidates rather than generating bounding box coordinates directly.

 \textbf{Instruction Tuning for Multi-Choice Region Selection.} 
While MLLMs are effective for cross-modality reasoning, which is crucial for REC, they may not perform well in the multi-choice region selection task. This is because they were not originally trained for this specific multiple-choice format. To address this, we propose instruction tuning tailored for multi-choice region selection. However, directly tuning the MLLM on the target dataset risks information leakage. Instead, we suggest that the MLLM only needs to acquire the ability to perform region selection, so we construct the instruction tuning dataset based on the external RefCOCO series, a standard REC benchmark that most MLLMs have already encountered during training.

Specifically, we generate a set of top-k bounding boxes for each referring expression using Grounding DINO and retain only those samples where the correct prediction (IoU $>$ 0.5) is within the top-k predictions. These are then shuffled to create candidate choices. This approach allows the MLLM to learn to distinguish between plausible bounding boxes and refine its ability to select the correct one based on both visual and linguistic context.

This instruction tuning strategy equips the MLLM to better handle the multiple-choice format, improving its region selection accuracy while maintaining computational efficiency.

 \textbf{Multi-Choice Instruction Tuning with \textit{None} Option.} 
 A key limitation of MLLMs is their difficulty in rejecting inappropriate queries, largely due to the lack of negative samples in the training data and well-defined refusal mechanisms. To address this, we propose an enhanced instruction tuning approach, sampling 10,000 positive and 2,500 negative samples from the FineCops-ref training set.

Expanding on the standard multiple-choice formulation, we introduce an additional ``None'' option (e.g., ``D. None'') to represent cases where no valid target exists. Our experimental results show that this approach significantly improves the model's ability to reject negative samples while maintaining high accuracy on positive samples.

\section{Experiments}
In this section, we present the experimental results. In subsection~\ref{evaluation}, we benchmark FineCops-Ref dataset by evaluating various representative models. The evaluation focuses on the models' ability to accurately localize the target and reject non-existent objects. In Subsection~\ref{method}, we assess the effectiveness of the proposed Specialist-MLLM collaboration approaches—namely, Slow-Fast Adaptation (SFA) and Candidate Region Selection (CRS)—through comparisons with existing state-of-the-art (SOTA) methods. Finally, in Subsection~\ref{ablation}, we provide the results of ablation studies to further validate our approach.

\subsection{Benchmarking the FineCops-Ref Dataset}
\label{evaluation}
\subsubsection{\textbf{Evaluated Models}}
We evaluate several representative models, including both Specialist Models and MLLMs. The models examined in this study include UNINEXT~\cite{yan2023universal}, SimVG~\cite{dai2024simvg}, MDETR~\cite{kamath2021mdetr}, MM-GDINO~\cite{zhao2024open,liu2023grounding},  Shikra~\cite{chen2023shikra}, Ferret~\cite{you2024ferret}, GroundingGPT~\cite{li-etal-2024-groundinggpt}, Lenna~\cite{wei2023lenna}, CogVLM~\cite{wang2023cogvlm}, CogCom~\cite{qi2024cogcom}, InternVL2.5~\cite{chen2024expanding} and Qwen2-vl~\cite{wang2024qwen2}. We use their open-source checkpoints for evaluation.
Additionally, we evaluate GPT4-V~\cite{achiam2023gpt}, which has limited capability in directly outputting bounding boxes. To assess its performance, we combine GPT4-V with the Set-of-Mark (SoM)~\cite{yang2023set}. Details on the model source and implementation are provided in the Appendix.

% We also test the effectiveness of training with positive samples from the training set of
% FineCops-Ref. We fine-tuned MM-GDINO-T, CogVLM, InternVL2.5 and Qwen2-vl using the positive data from FineCops-Ref. In addition, we fine-tuned MM-GDINO-T with the entire training set. The training settings are detailed in Appendix~\ref{sec:appendix_exp}.

\begin{table}
\centering
\caption{Evaluation results (Precision@1) on positive data. L1, L2, and L3 represent Level1, Level2, and Level3 of the ground truth labels in FineCops-Ref, respectively. The best results are in bold, and the second-best results are underlined.}
\label{precision}
\resizebox{\columnwidth}{!}{
\begin{tabular}{lcccc} 
\toprule
                                     & \multicolumn{4}{c}{\textbf{Positive}}  \\ 
\cmidrule(lr){2-5}
\textbf{\textbf{Model}}              & L1    & L2    & L3    & \textbf{All}   \\ 
\midrule
\textbf{Specialist}                  &       &       &       &                \\
UNINEXT~\cite{yan2023universal}                              & 59.95 & 43.60 & 40.98 & 53.22          \\
SimVG~\cite{dai2024simvg}                                & 61.88 & 47.30 & 42.04 & 55.74          \\
MM-GDINO-T~\cite{zhao2024open}                           & 75.11 & 34.78 & 35.46 & 58.87          \\
MDETR~\cite{kamath2021mdetr}                                & 73.25 & 53.08 & 46.71 & 64.80          \\
MM-GDINO-L~\cite{zhao2024open}                           & \textbf{85.13} & 43.54 & 42.89 & 68.32          \\
\midrule
\textbf{MLLM}                        &       &       &       &                \\
GPT4-V~\cite{achiam2023gpt} + SoM~\cite{yang2023set}                         & 55.94 & 45.94 & 49.29 & 52.07          \\
Shikra~\cite{chen2023shikra}                               & 64.64 & 50.29 & 43.95 & 58.54          \\
Lenna~\cite{wei2023lenna}                                & 73.75 & 41.92 & 38.43 & 60.74          \\
GroundingGPT~\cite{li-etal-2024-groundinggpt}                         & 71.10 & 53.64 & 49.89 & 63.87          \\
Ferret-13B~\cite{you2024ferret}                           & 70.24 & 55.99 & 50.53 & 64.23          \\
InternVL2.5-8B~\cite{chen2024expanding}                       & 70.91 & 59.67 & 53.08 & 66.05          \\
CogVLM~\cite{wang2023cogvlm}                               & 74.59 & \underline{62.51} & 57.32 & 69.46          \\
CogCom~\cite{qi2024cogcom}                               & 76.27 & 60.93 & \underline{59.87} & \underline{70.03}          \\
Qwen2-VL-7B~\cite{wang2024qwen2}                           & \underline{78.24} & \textbf{66.13} & \textbf{60.72} & \textbf{73.09}          \\
\bottomrule 		 	 
\end{tabular}}
\end{table} 	 

\subsubsection{\textbf{Evaluation on Positive Data}}
\label{sec:positive_eval }
We evaluate the models using Precision@1 for positive data, as summarized in Table~\ref{precision}. The results highlight the importance of categorizing the dataset by difficulty levels, as the performance of most models degrades significantly with increasing complexity. Notably, at difficulty level 3, the majority of models achieve a precision below 50\%.
An interesting observation is that models like UNINEXT, SimVG, and InternVL2.5, which achieve state-of-the-art performance on the RefCOCO series benchmarks, perform relatively poorly on the FineCops-Ref benchmark. In contrast, Grounding DINO and Qwen2-VL demonstrate strong generalization capabilities, achieving the best results in the Specialist and MLLM categories, respectively.

\textbf{Specialist performs better on simple REC tasks.}
At Level 1, the task primarily involves identifying objects based on category names, closely aligning with the requirements of open-vocabulary object detection. Grounding DINO, built on the SWIN-L backbone, achieves an impressive accuracy of 85.13\% under this settings. This observation leads to two key insights. First, vision-language models focused on object detection exhibit strong capabilities in basic visual localization and object detection tasks, even in zero-shot scenarios, which is also supported by their superior performance on RefCOCO benchmark which mainly require the model to detect the obejct without consider the intricate attribute and relation. Second, although MLLMs excel in language understanding and reasoning, their basic object detection abilities still fall short of the standards required for truly general-purpose models.

\textbf{MLLMs exhibit superior reasoning abilities.}
At Level 2 and 3, the tasks demand sophisticated language understanding and cross-modal reasoning due to the presence of numerous easily confusable objects in the images. However, most models do not exhibit sufficient capability in this regard. MLLMs based on large language models (LLMs) perform better in this aspect, showcasing their superior compositional reasoning abilities.

\subsubsection{\textbf{Evaluation on Negative Data}}
We evaluate the models using Recall@1 and AUROC for negative data. The AUROC results can be found in Appendix. Specifically, models like MDETR and Lenna that have dedicated object detection modules can generate multiple detection boxes with associated confidence scores, allowing for direct computation of Recall@1 and AUROC. For models that generate coordinates as text using an autoregressive approach, we use the probability of the coordinate tokens to calculate confidence~\cite{kurita2023refego,mitchell2023detectgpt}.
The evaluation results for negative expressions and negative images are shown in Table~\ref{recall_text} and Table~\ref{recall_image}, respectively. The results indicate that existing models generally exhibit low recall@1 performance on negative data, with most values falling below 50\%. This suggests that these models have weak rejection capability for non-existent objects. In contrast, the latest model, Qwen2-VL, demonstrates strong generalization ability and achieves the best performance.
According to the experimental results, we can draw the following conclusions:

\begin{table*}[ht]
\centering
\caption{Evaluation results (Recall@1) on negative expressions. L1 and L2 denote the direct changes related to the target and the changes related to other objects within the expression, respectively.}
\label{recall_text}
\resizebox{0.75\textwidth}{!}{
\begin{tabular}{lccccccccccc} 
\toprule
                                     & \multicolumn{6}{c}{\textbf{REPLACE}}                                                                                 & \multicolumn{4}{c}{\textbf{SWAP}}                                            &                \\ 
\cmidrule(r){2-7}\cmidrule(r){8-11}
                                     & \multicolumn{2}{c}{\textbf{Object}} & \multicolumn{2}{c}{\textbf{Attribute}} & \multicolumn{2}{c}{\textbf{Relation}} & \multicolumn{2}{c}{\textbf{Object}} & \multicolumn{2}{c}{\textbf{Attribute}} &   \\ 
\cmidrule(lr){2-3}\cmidrule(r){4-5}\cmidrule(r){6-7}\cmidrule(lr){8-9}\cmidrule(lr){10-11}
\textbf{Model}                       & L1    & L2                          & L1    & L2                             & L1    & L2                            & L1    & L2                          & L1    & L2                             & \textbf{Avg.}  \\ 
\midrule
\textbf{Specialist}                  &       &                             &       &                                &       &                               &       &                             &       &                                &                \\
MM-GDINO-T~\cite{zhao2024open}                           & 58.84 & 33.77                       & 50.47 & 29.96                          & 34.69 & 31.92                         & 43.89 & 27.71                       & 43.67 & 31.97                          & 38.69          \\
UNINEXT~\cite{yan2023universal}                              & 47.83 & 33.70                       & 44.66 & 34.30                          & 39.51 & 35.61                         & 45.31 & 37.35                       & 41.69 & 31.97                          & 39.19          \\
MDETR~\cite{kamath2021mdetr}                                & 52.89 & 36.09                       & 50.47 & 35.92                          & 42.48 & 40.77                         & 45.89 & 37.35                       & 44.42 & 37.70                          & 42.40          \\
SimVG~\cite{dai2024simvg}       & 43.45 & 43.20 & 46.46 & 44.04 & 40.33 & \textbf{45.20} & 41.00 & 31.93 & 40.20 & \textbf{50.82} & 42.66          \\
MM-GDINO-L~\cite{zhao2024open}                           & \underline{64.23} & 40.26                       & 55.76 & 41.52                          & 45.74 & 43.73                         & \textbf{53.02} & \underline{48.19}                       & \underline{49.38} & 37.70                          & 47.95          \\
\midrule
\textbf{MLLM}                        &       &                             &       &                                &       &                               &       &                             &       &                                &                \\
Ferret-13B~\cite{you2024ferret}                           & 38.38 & 33.01                       & 37.57 & 34.48                          & 35.58 & 34.69                         & 38.69 & 34.94                       & 35.73 & 35.25                          & 35.83          \\
GroundingGPT~\cite{li-etal-2024-groundinggpt}                         & 42.24 & 35.13                       & 40.14 & 33.75                          & 37.51 & 36.72                         & 41.77 & 39.76                       & 35.24 & 39.34                          & 38.16          \\
Shikra~\cite{chen2023shikra}                               & 44.99 & 33.11                       & 41.25 & 33.03                          & 35.78 & 39.85                         & 42.27 & 39.16                       & 39.70 & 32.79                          & 38.19          \\
InternVL2.5-8B~\cite{chen2024expanding}                       & 38.55 & 36.16                       & 41.84 & 37.91                          & 40.92 & 39.48                         & 39.40 & 39.16                       & 43.92 & 38.52                          & 39.58          \\
CogVLM~\cite{wang2023cogvlm}                               & 53.34 & 44.02                       & 51.24 & \textbf{48.74}                          & 41.22 & \underline{44.46}                         & 47.69 & \textbf{49.40}                       & 46.40 & 40.16                          & 46.67          \\
CogCom~\cite{qi2024cogcom}                               & 57.96 & 44.91                       & 54.65 & 44.04                          & 45.81 & 41.70                         & \underline{51.03} & 43.98                       & 47.39 & 36.89                          & 46.84          \\
Lenna~\cite{wei2023lenna}                                & \textbf{65.88} & \textbf{50.38}                       & \underline{58.75} & 42.96                          & \underline{47.00} & 43.91                         & 49.94 & 38.55                       & \underline{49.38} & 43.44                          & \underline{49.02}          \\
Qwen2-VL-7B~\cite{wang2024qwen2}                           & 60.63 & \underline{45.62}                       & \textbf{59.59} & \underline{46.93}                          & \textbf{49.18} & 42.44                         & 49.13 & 46.99                       & \textbf{50.87} & \underline{44.26}                          & \textbf{49.56}          \\
\bottomrule
\end{tabular}}
\end{table*}

\begin{table*}[!ht]
\centering
\caption{Evaluation results (Recall@1) on negative images.}
\label{recall_image}
\resizebox{0.75\textwidth}{!}{
\begin{tabular}{lcccccccccc} 
\toprule
                                     & \multicolumn{4}{c}{\textbf{REPLACE}}                                         & \multicolumn{5}{c}{\textbf{SWAP}}                                                            &                \\ 
\cmidrule(lr){2-5}\cmidrule(lr){6-10}
                                     & \multicolumn{2}{c}{\textbf{Object}} & \multicolumn{2}{c}{\textbf{Attribute}} & \textbf{Object} & \multicolumn{2}{c}{\textbf{Attribute}} & \multicolumn{2}{c}{\textbf{Flip}} &                \\ 
\cmidrule(lr){2-3}\cmidrule(lr){4-5}\cmidrule(lr){6-6}\cmidrule(lr){7-8}\cmidrule(lr){9-10}
\textbf{Model}                       & L1    & L2                          & L1    & L2                             & L1              & L1    & L2                             & L1    & L2                        & \textbf{Avg.}  \\ 
\midrule
\textbf{Specialist}                  &       &                             &       &                                &                 &       &                                &       &                           &                \\
UNINEXT~\cite{yan2023universal}      & 48.85 & 31.62                       & 40.96 & 30.33                          & 46.91           & 40.25 & 37.06                          & 30.66 & 29.42                     & 37.34          \\
SimVG~\cite{dai2024simvg}            & 41.69 & 41.50                       & 43.73 & 43.15                          & 31.60           & 38.60 & \underline{51.75}               & 36.74 & 39.66                     & 40.94          \\
MM-GDINO-T~\cite{zhao2024open}       & 58.46 & 40.73                       & 44.75 & 37.61                          & 46.25           & 51.33 & 28.67                          & 39.50 & 40.94                     & 43.14          \\
MDETR~\cite{kamath2021mdetr}         & 58.15 & 42.85                       & 51.70 & 37.95                          & 48.86           & 49.49 & 44.76                          & 44.29 & 42.22                     & 46.70          \\
MM-GDINO-L~\cite{zhao2024open}       & 66.35 & 49.45                       & \underline{54.93} & \underline{49.05}               & \underline{55.05} & \textbf{62.63} & 46.85                          & \underline{45.21} & \underline{46.48}          & \underline{52.89} \\ 
\midrule
\textbf{MLLM}                        &       &                             &       &                                &                 &       &                                &       &                           &                \\
CogCom~\cite{qi2024cogcom}           & 32.24 & 21.55                       & 22.57 & 20.10                          & 39.74           & 25.46 & 18.88                          & 24.13 & 23.03                     & 25.30          \\
Shikra~\cite{chen2023shikra}         & 42.57 & 33.61                       & 36.54 & 34.26                          & 35.18           & 38.60 & 36.36                          & 34.25 & 37.10                     & 36.50          \\
GroundingGPT~\cite{li-etal-2024-groundinggpt} & 43.91 & 36.88                       & 36.31 & 35.88                          & 39.09           & 37.17 & 40.56                          & 37.02 & 33.05                     & 37.76          \\
Ferret-13B~\cite{you2024ferret}      & 41.54 & 37.46                       & 38.04 & 36.22                          & 43.00           & 37.78 & 39.16                          & 35.27 & 36.25                     & 38.30          \\
InternVL2.5-8B~\cite{chen2024expanding} & 42.96 & 41.57                       & 39.46 & 40.80                          & 45.93           & 44.35 & 45.45                          & 42.08 & 40.72                     & 42.59          \\
Lenna~\cite{wei2023lenna}            & \underline{66.88} & \underline{51.19}           & 54.38 & 39.34                          & 47.56           & 49.08 & 43.36                          & 33.98 & 30.92                     & 46.30          \\
CogVLM~\cite{wang2023cogvlm}         & 51.11 & 49.01                       & 43.49 & 46.10                          & 50.49           & \underline{53.80} & 49.65                          & 43.74 & 37.74                     & 47.24          \\
Qwen2-VL-7B~\cite{wang2024qwen2}     & \textbf{69.73} & \textbf{60.65}               & \textbf{63.87} & \textbf{58.41}                  & \textbf{63.52} & \textbf{62.63} & \textbf{54.55}                  & \textbf{61.88} & \textbf{52.45}             & \textbf{60.85} \\ 
\bottomrule
\end{tabular}}
\end{table*}

% \textbf{The models exhibit significant sensitivity to different replacement content.}
\textbf{The models struggle with compositional reasoning and exhibit a limited understanding of semantic structures.}
L1 and L2 represent two levels of modification: L1 refers to direct changes related to the target, including objects, attributes, and relations, while L2 involves changing other object-related elements, which are often more challenging to detect.
For the same type of negative data, the recall of L1 is typically significantly higher than that of L2, suggesting that most models can effectively identify simple anomalies, such as changes directly related to the target. However, for L2 negative data, almost all models perform poorly. This further highlights that the models lack compositional reasoning ability and fail to fully comprehend the semantic structure of sentences.

\textbf{The models exhibit poor understanding of fine-grained attributes and relations.}
For negative data at the same level, the recall of changing attributes is generally lower than that of changing objects, while the recall of changing relations is typically significantly lower than the first two. This suggests that the models exhibit relatively stronger object recognition capabilities but weaker understanding of attributes and relations. Specifically, the models perform poorly on negative data involving the ``replace relation'' and ``flip'' types, indicating substantial challenges in relation comprehension. Another notable observation is that, compared to directly replacing attributes, the models perform worse when identifying negative data of the ``swap attribute'' type, highlighting their limitations in accurately associating attributes.

% While models demonstrate relatively strong recognition capabilities for direct object replacement—cases where the target mentioned in the expression is completely absent from the image—their ability to recognize attributes is notably weaker. Furthermore, the models face significant challenges in understanding relationships, particularly in recognizing replaced relationships and altered word order, consistent with findings from previous studies. An additional observation is that the models perform worse in identifying negative data of the ``swap attribute'' type compared to direct attribute replacements, highlighting deficiencies in their ability to accurately bind attributes.

% \subsubsection{Recall and AUROC for Option}
% \begin{itemize}
%     \item Calculate metrics based on answer confidence.
%     \item Calculate based on confidence of 'None' answers.
% \end{itemize}

% \begin{itemize}
%     \item Discuss how adding a "None" option can improve rejection capability.
%     \item Experiment with the inclusion of a "None" option.
% \end{itemize}

% \subsection{Other Benchmark Testing}
% \begin{itemize}
%     \item Evaluate the model on other benchmarks.
%     \item Consider reviewer comments in the evaluation.
% \end{itemize}

\begin{table*}[ht]
\centering
\caption{Comparison with state-of-the-art Specialist and MLLM methods on three REC benchmarks. L1 and L2 denote Level 1 and Level 2, respectively, as determined by the router Grounding DINO~\cite{zhao2024open} in SFA. $\dagger$ indicates the use of the focus-enhancement strategy in our SFA method.}
\label{SFA}
\resizebox{0.9\textwidth}{!}{
\begin{tabular}{lccccccccc}
\toprule
                       & \multicolumn{3}{c}{\textbf{FineCops-Ref}} & \multicolumn{3}{c}{\textbf{Ref-Adv}}  & \multicolumn{3}{c}{\textbf{Ref-Reasoning}}  \\ 
\cmidrule(l){2-10}
\textbf{Method}        & L1    & L2    & \textbf{All}              & L1    & L2    & \textbf{All}          & L1    & L2    & \textbf{All}                \\ 
\midrule
\textbf{Specialist}    &       &       &                           &       &       &                       &       &       &                             \\
UNINEXT~               & 65.39 & 39.06 & 53.22                     & 82.55 & 69.68 & 75.59                 & 35.03 & 16.66 & 20.80                       \\
SimVG~                 & 66.86 & 42.80 & 55.74                     & 82.90 & 69.83 & 75.84                 & 35.79 & 18.64 & 22.50                       \\
MDETR~                 & 74.85 & 53.10 & 64.80                     & 81.20 & 67.38 & 73.73                 & 46.53 & 26.62 & 31.10                       \\
MM-GDINO-L~            & 81.20 & 53.32 & 68.32                     & 85.37 & 66.08 & 74.95                 & 50.94 & 27.09 & 32.46                       \\ 
\midrule
\textbf{MLLM}          &       &       &                           &       &       &                       &       &       &                             \\
Shikra~                & 66.92 & 48.79 & 58.54                     & 79.64 & 66.62 & 72.31                 & 40.85 & 23.83 & 27.18                       \\
GroundingGPT~          & 75.38 & 50.48 & 63.87                     & 80.20 & 66.68 & 72.89                 & 43.30 & 21.69 & 26.55                       \\
Ferret-13B~            & 73.77 & 53.12 & 64.23                     & 81.61 & 71.73 & 76.27                 & 42.27 & 23.23 & 27.51                       \\
InternVL2.5-8B~        & 76.05 & 54.40 & 66.05                     & 80.85 & 71.73 & 75.92                 & 44.44 & 25.97 & 30.13                       \\
CogVLM~                & 73.58 & 64.68 & 69.46                     & 79.14 & 71.98 & 75.27                 & 45.67 & 33.10 & 35.93                       \\
CogCom~                & 77.89 & 60.87 & 70.03                     & 82.20 & 71.23 & 76.27                 & 47.29 & 28.44 & 32.68                       \\
Qwen2-VL-7B~           & 79.40 & 65.74 & 73.09                     & 85.84 & 74.73 & 79.83                 & 50.62 & 31.46 & 35.77                       \\ 
\midrule
% \textbf{SFA}           & 81.20 & 65.74 & \textbf{74.06(+0.97)}     & 85.37 & 74.73 & \textbf{79.62(-0.21)} & 50.94 & 33.10 & \textbf{37.11(+1.34)}       \\
% \textbf{SFA$\dagger$~} & 83.43 & 66.91 & \textbf{75.61(+2.52)}     & 84.78 & 75.92 & \textbf{79.97(+0.14)} & 52.34 & 34.37 & \textbf{38.41(+2.48)}       \\
\textbf{SFA}           &       &       &                           &       &       &                       &       &       &                             \\
CogVLM(SFA)           & 81.20 & 64.68 & \textbf{73.57(+4.11)}     & 85.37 & 71.98 & \textbf{78.13(+2.86)} & 50.94 & 33.10 & \textbf{37.11(+1.18)}       \\
CogVLM(SFA$\dagger$)           & 83.43 & 65.69 & \textbf{75.03(+5.57)}     & 85.37 & 72.98 & \textbf{78.37(+3.10)} & 52.34 & 34.37 & \textbf{38.41(+2.48)}       \\
Qwen2-VL-7B(SFA)           & 81.20 & 65.74 & \textbf{74.06(+0.97)}     & 85.37 & 74.73 & \textbf{79.62(-0.21)} & 50.94 & 31.46 & \textbf{35.84(+0.07)}       \\
Qwen2-VL-7B(SFA$\dagger$)           & 83.43 & 66.91 & \textbf{75.61(+2.52)}     & 84.78 & 75.92 & \textbf{79.97(+0.14)} & 52.34 & 31.72 & \textbf{36.36(+0.59)}       \\
\midrule
\textbf{CRS}           &       &       &                           &       &       &                       &       &       &                             \\
CogVLM(CRS)            & 79.13 & 67.20 & \textbf{73.62(+4.16)}     & 81.37 & 74.58 & \textbf{77.70(+2.43)} & 51.27 & 33.97 & \textbf{37.87(+1.94)}       \\
Qwen2-VL-7B(CRS)       & 84.77 & 70.85 & \textbf{78.33(+5.24)}     & 84.37 & 77.32 & \textbf{80.56(+0.73)} & 54.80 & 36.28 & \textbf{40.45(+4.68)}       \\
InternVL2.5-8B(CRS)    & 86.06 & 69.88 & \textbf{78.58(+12.53)}    & 87.96 & 78.42 & \textbf{82.80(+6.88)} & 55.10 & 36.05 & \textbf{40.34(+10.21)}      \\
\bottomrule
\end{tabular}}
\end{table*}

\begin{figure*}[h]
  \centering
  \includegraphics[width=0.88\textwidth]{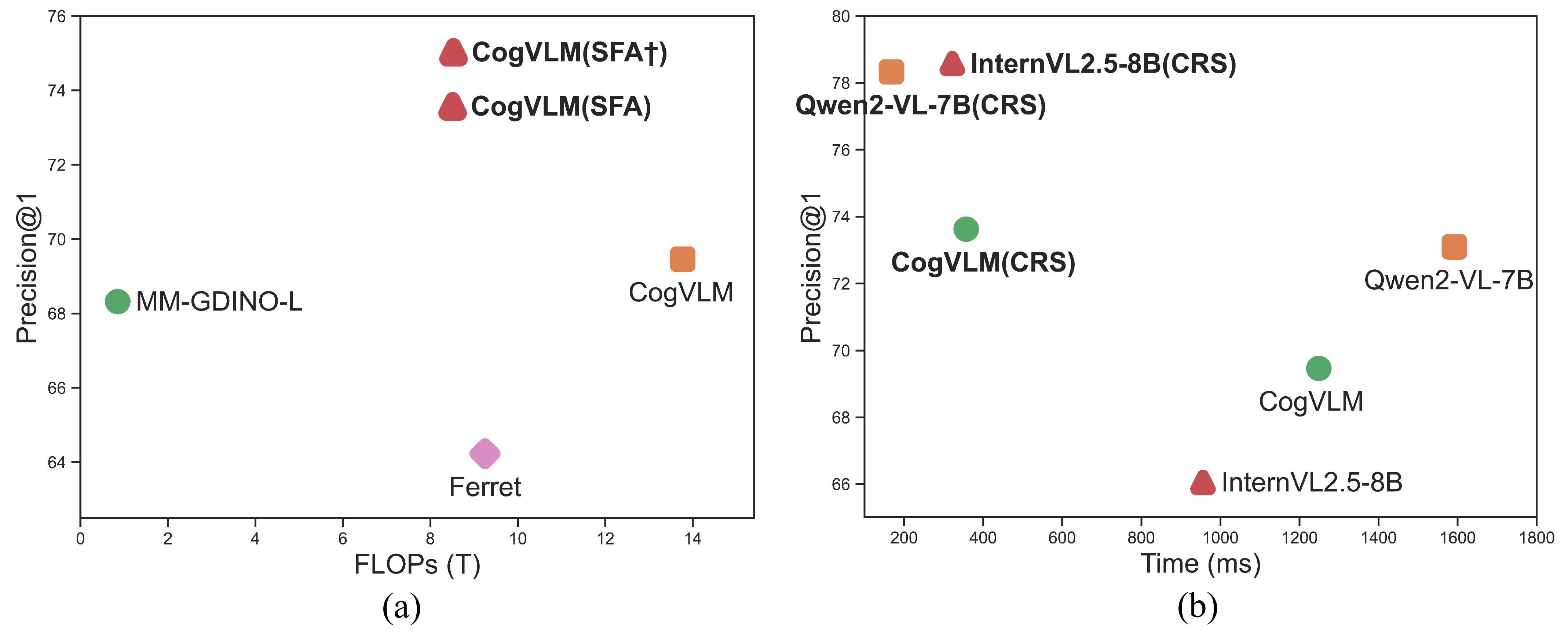}
  \caption{The effectiveness of the proposed methods. (a) compares the FLOPs (computational cost) of SFA with other methods, including the Specialist Model MM-GDINO-L~\cite{zhao2024open} and MLLMs such as CogVLM~\cite{wang2023cogvlm} and Ferret~\cite{you2024ferret}. (b) compares the inference time of CRS with that of MLLMs themselves.}
  \label{effectiveness}
\end{figure*}

\subsection{Evaluation of the Proposed Specialist-MLLM Collaboration Approaches}
\label{method}
In this section, we compare our proposed Specialist-MLLM collaboration methods with representative REC models, including both Specialist Models and MLLMs. We evaluate the performance using Precision@1. The details for CRS training are provided in Appendix.

\subsubsection{\textbf{Evaluation of Slow-Fast Adaptation (SFA)}}
As shown in Table~\ref{SFA}, Specialist Models generally perform well on Level 1 tasks, where text-to-region feature matching is essential, while MLLM solutions tend to excel in Level 2 tasks, which require more complex reasoning. This observation supports our task-adaptive model selection strategy. Our SFA approach, which combines the strengths of both methods through effective task routing, achieves superior performance compared to using these strategies individually. Incorporating the focus-enhancement strategy further boosts performance, with our SFA$\dagger$ yielding additional improvements. These results demonstrate the effectiveness of our solution in handling REC tasks with varying levels of difficulty, highlighting the broad applicability of our approach.

Importantly, thanks to task-specific model selection, our method achieves high efficiency compared to relying solely on MLLMs. As illustrated in Fig.~\ref{effectiveness}(a), our approach outperforms CogVLM~\cite{wang2023cogvlm} by an absolute 5.6\% while using only 62\% of the computational cost. When compared to Ferret~\cite{you2024ferret}, which maintains similar computational overhead, our method achieves over a 10\% performance boost. These results demonstrate that our method strikes an optimal balance between performance and efficiency.

\textbf{Effectiveness of Target-refocus Strategy.}
To evaluate the generalizability of our proposed focus-enhancement strategy, we apply it to various Specialist and MLLM methods, as shown in Table~\ref{focus}. This strategy helps models focus on the referred target, thereby reducing the misidentification of the target object as another category. The experimental results show that our focus-enhancement strategy consistently improves the performance of both Specialist and MLLM methods across all datasets, demonstrating its effectiveness and broad applicability.

\begin{table}
\centering
\caption{Effectiveness of the proposed focus-enhancement strategy. \textit{Exp.} denotes the original method, \textit{Foc.} indicates our focus-enhancement strategy, which can be readily applied to various Specialist and MLLM methods.}
\label{focus}
\resizebox{0.5\textwidth}{!}{
\begin{tabular}{lcccc} 
\toprule
\textbf{Model}                  & \textbf{Method} & \textbf{FineCops-Ref} & \textbf{Ref-Adv} & \textbf{Ref-Reasoning}  \\ 
\midrule
\textbf{Specialist}             &                 &                       &                  &                         \\ 
\cmidrule(r){1-5}
\multirow{2}{*}{MDETR~\cite{kamath2021mdetr}}          & Exp.            & 64.80                 & 73.73            & 31.10                   \\
                                & Foc.            & 67.19(+2.39)         & 72.12(-1.61)    & 34.65(+3.55)           \\ 
\cmidrule(r){1-5}
\multirow{2}{*}{MM-GDINO-L~\cite{zhao2024open}}     & Exp.            & 68.32                 & 74.95            & 32.46                   \\
                                & Foc.            & 71.22(+2.90)         & 74.55(-0.40)    & 34.28(+1.82)           \\ 
\midrule
\textbf{MLLM}                   &                 &                       &                  &                         \\ 
\cmidrule(r){1-5}
\multirow{2}{*}{GroundingGPT~\cite{li-etal-2024-groundinggpt}}   & Exp.            & 63.87                 & 72.89            & 26.55                   \\
                                & Foc.            & 65.90(+2.03)         & 74.33(+1.44)    & 28.11(+1.56)           \\ 
\cmidrule(lr){1-5}
\multirow{2}{*}{Ferret-13B~\cite{you2024ferret}}     & Exp.            & 64.23                 & 76.27            & 27.51                   \\
                                & Foc.            & 66.10(+1.87)         & 78.07(+1.80)    & 28.96(+1.45)           \\ 
\cmidrule(r){1-5}
\multirow{2}{*}{CogVLM~\cite{wang2023cogvlm}}         & Exp.            & 69.46                 & 75.27            & 35.93                   \\
                                & Foc.            & 72.11(+2.65)         & 77.75(+2.48)    & 37.12(+1.19)           \\ 
\cmidrule(lr){1-5}
\multirow{2}{*}{InternVL2.5-8B~\cite{chen2024expanding}} & Exp.            & 66.05                 & 75.92            & 30.13                   \\
                                & Foc.            & 69.89(+3.84)         & 80.29(+4.37)    & 32.29(+2.16)           \\ 
\cmidrule(lr){1-5}
\multirow{2}{*}{Qwen2-VL-7B~\cite{wang2024qwen2}}     & Exp.            & 73.09                 & 79.83            & 35.77                   \\
                                & Foc.            & 75.03(+1.94)         & 80.89(+1.06)    & 36.16(+0.39)           \\
\bottomrule
\end{tabular}}
\end{table}

\begin{table}[h]
\centering
\caption{Evaluation of CRS effectiveness in non-existent target rejection across three settings: (1) CRS without instruction tuning (w/o instruction tuning); (2) instruction tuning with the original prompt (baseline); and (3) instruction tuning with the CRS prompt (CRS).}
\label{neg_opt}
\resizebox{0.5\textwidth}{!}{
\begin{tabular}{llcccc} 
\toprule
                              &                 & \multicolumn{2}{c}{\textbf{Accuracy}} & \multicolumn{2}{c}{\textbf{Recall}}  \\ 
\cmidrule(lr){3-4}\cmidrule(lr){5-6}
\textbf{Finetune}             & \textbf{Model}  & positive & negative                   & text  & image                        \\ 
\midrule
\multirow{2}{*}{w/o instruction tuning} & InternVL2.5-8B~ & 73.24    & 11.80                      & 54.91 & 54.69                        \\
                              & Qwen2-VL-7B~    & 74.50    & 5.20                       & 55.37 & 56.59                        \\ 
\midrule
\multirow{3}{*}{Baseline}     & CogVLM          & 67.83    & 0.99                       & 45.76 & 45.84                        \\
                              & Internvl2.5-8B  & 69.90    & 58.32                      & 55.06 & 56.00                        \\
                              & Qwen2vl-7B      & 64.45    & 62.31                      & 55.30 & 54.97                        \\ 
\midrule
\multirow{3}{*}{CRS}          & CogVLM~         & 70.57    & 36.52                      & 53.65 & 55.48                        \\
                              & InternVL2.5-8B  & 76.74    & 58.63                      & 69.54 & 68.31                        \\
                              & Qwen2-VL-7B     & 79.00    & 27.08                      & 64.98 & 68.16                        \\
\bottomrule
\end{tabular}}
\end{table}

\subsubsection{\textbf{Evaluation of Candidate Region Selection (CRS)}}
As shown in Table~\ref{SFA}, our CRS method consistently achieves improvements across all three benchmarks. Notably, when applied to InternVL2.5~\cite{chen2024expanding}, CRS delivers a remarkable performance boost of nearly 10\%. The experimental results demonstrate that advanced models possess substantial cross-modal reasoning capability. Through instruction tuning, we effectively harness these inherent capabilities, enabling the model to achieve stronger grounding performance and fully unleash its potential. These findings underscore both the effectiveness and generalizability of our approach.

\begin{figure}
  \centering
  \includegraphics[width=1\linewidth]{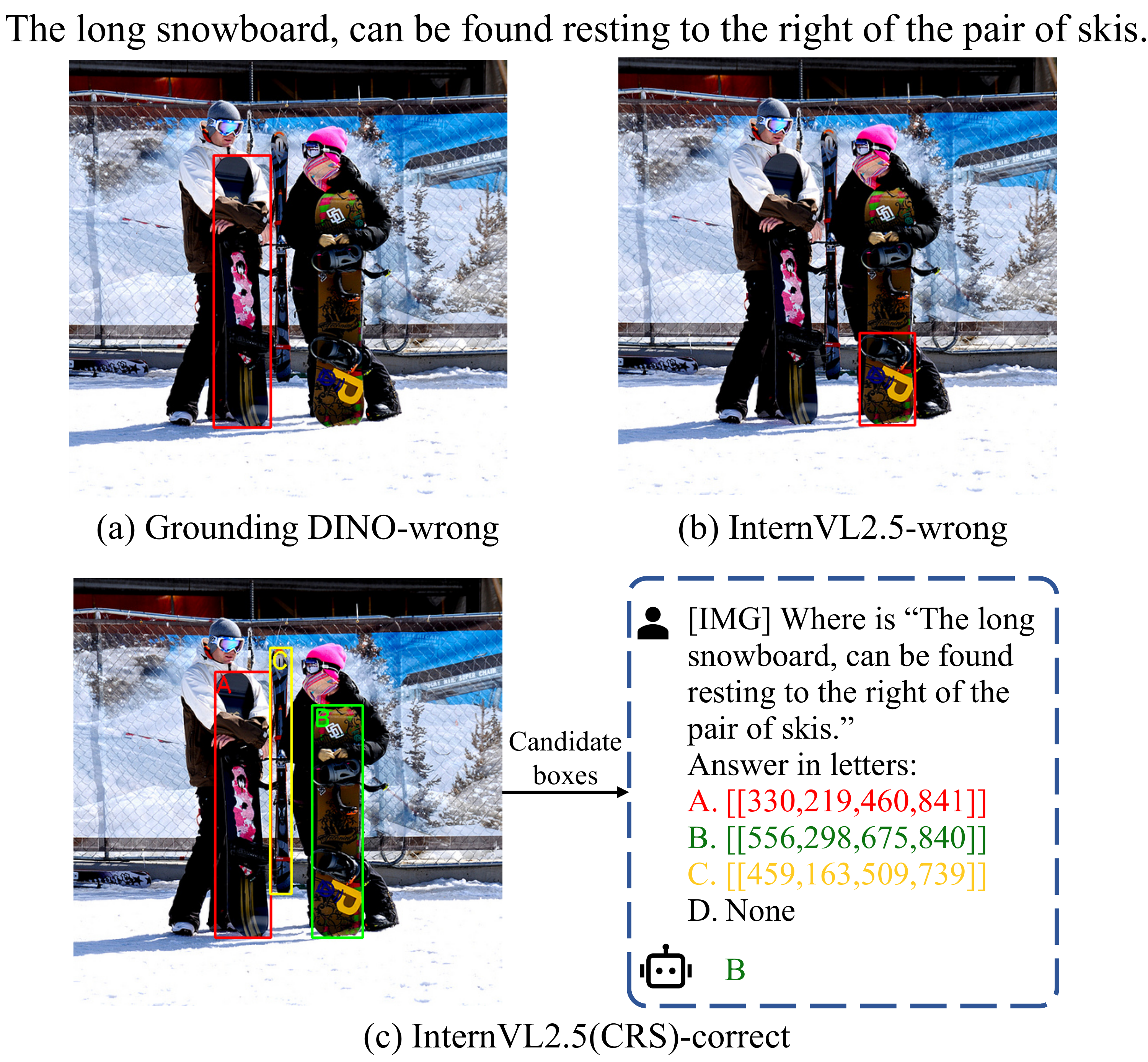}
  \caption{Visualization of grounding results for Grounding DINO~\cite{zhao2024open}, InternVL2.5~\cite{chen2024expanding}, and the CRS approach.}
  \label{crs}
\end{figure}

It is important to note that, while CRS slightly increases the length of input tokens, it significantly reduces the length of output tokens (from n to 1). In practical applications, this reduction leads to a considerable improvement in inference speed. As illustrated in Fig.~\ref{effectiveness}(b), CRS requires only 10\%-30\% of the inference time compared to the original method, underscoring the strong practical value of our approach.

As shown in Fig.~\ref{crs}, both Grounding DINO and InternVL2.5 output wrong bounding boxes. However, after applying the CRS method, InternVL2.5, leveraging the candidate boxes provided by Grounding DINO and effectively comprehending the relationship in the expression, successfully selects the correct answer: ``B''.

% \begin{table*}
% \centering
% \caption{Effectiveness of the proposed Candidate Region Selection. \textit{Ori.} denotes the original method, \textit{Sel.} indicates Candidate Region Selection. The best results are highlighted in bold.}
% \label{select}
% \resizebox{0.8\textwidth}{!}{
% \begin{tabular}{lccccc} 
% \toprule
% \textbf{Model}                  & \textbf{Method} & \textbf{FineCops-Ref} & \textbf{Ref-Adv} & \textbf{Ref-Reasoning} & \textbf{Ref-L4}  \\ 
% \midrule
% \multirow{2}{*}{CogVLM}         & Ori.            & 69.46                 & 75.27            & 35.93                  & 82.68            \\
%                                 & Sel.            & 73.62                 & 77.70            & 37.87                  & 82.57            \\ 
% \midrule
% \multirow{2}{*}{Qwen2-VL-7B}     & Ori.            & 73.09                 & 79.83            & 35.77                  & 78.68            \\
%                                 & Sel.            & 77.84                 & 80.16            & 40.24                  & 81.51            \\ 
% \midrule
% \multirow{2}{*}{InternVL2.5-8B} & Ori.            & 66.05                 & 75.92            & 30.13                  & 76.62            \\
%                                 & Sel.            & 78.58                 & 82.80            & 40.34                  & 84.09            \\
% \bottomrule
% \end{tabular}}
% \end{table*}

Another advantage of our CRS strategy is its inherent ability to reject non-existent objects by selecting a ``None'' option. To evaluate this capability, we assess three approaches: (1) using the CRS method without instruction tuning (w/o instruction tuning); (2) instruction tuning an MLLM model with its original REC prompt, followed by the instruction ``If no target object exists, please output ``None'' (Baseline); and (3) instruction tuning the model using the CRS prompt, followed by the instruction ``If no suitable option exists, please select the option corresponding to ``None'' (CRS).
% Using the region selection method, we append a ``None'' option to the end of the choices and prompt the model with: ``If no suitable option exists, please select the option corresponding to `None'.'' This evaluation approach enables the model to explicitly reject negative data, providing an additional reference metric. Without fine-tuning, we evaluate two of the most generalizable models currently available: InternVL2.5 and Qwen2-VL. Additionally, we fine-tune three advanced models—CogVLM, InternVL2.5, and Qwen2-VL—using data constructed from the FineCops-Ref training set, which contains negative samples where the model is expected to select the ``None'' option. The training configurations are detailed in the Appendix~\ref{sec:appendix_results}.
We evaluate the models using Accuracy and Recall@1, as summarized in Table~\ref{neg_opt}. For positive data, Accuracy is equivalent to Precision@1, while for negative data, Accuracy measures the proportion of negative data for which the model correctly selects the ``None'' option. Recall@1 reflects the average recall score.

The experimental results indicate that, without instruction tuning, models frequently fail to select ``None,'' suggesting that existing models have limited rejection capabilities for non-existent objects. This observation aligns with our previous conclusions. After instruction tuning with the baseline method, models manage to respond with ``None,'' but their accuracy on positive data drops significantly. This indicates that models fail to properly learn when they should reject. In contrast, instruction tuning with the CRS method improves accuracy on both positive and negative data and demonstrates a substantial advantage in Recall@1. These results validate the effectiveness of the CRS method: the models successfully learn to reject non-existent objects while maintaining precise localization performance for positive data.

\begin{table*}[ht]
\centering
\caption{Ablation study of different detection-based difficulty-level assessment methods. Difficulty-level assessment accuracy and L2 recall are evaluated on the FineCops-Ref benchmark, based on ground-truth labels provided.}
\label{ablation_router}
\resizebox{0.8\textwidth}{!}{
\begin{tabular}{lcccccc} 
\toprule
\textbf{Method} & \textbf{Accuracy} & \textbf{L2 Recall} & \textbf{FineCops-Ref} & \textbf{Ref-Adv} & \textbf{Ref-Reasoning} & \begin{tabular}[c]{@{}c@{}}\textbf{Time}\\\textbf{(ms)}\end{tabular}  \\ 
\midrule
OV-DINO~\cite{wang2024ov}        & 58.46             & 34.99              & 70.39                 & 77.11            & 35.32                  & 112                                                                   \\
ProxyDet~\cite{jeong2024proxydet}       & 57.82             & 46.53              & 70.97                 & 78.21            & 34.76                  & 79                                                                    \\
YOLO-World~\cite{cheng2024yolo}     & 67.35             & 28.70              & 71.53                 & 77.81            & 34.15                  & 44                                                                    \\
MDETR~\cite{kamath2021mdetr}          & 61.64             & 63.34              & 72.72                 & 79.29            & 36.18                  & \textbf{26}                                                                    \\
MM-GDINO-L~\cite{zhao2024open}     & \textbf{66.75}             & \textbf{66.06}              & \textbf{74.06}                 & \textbf{79.62}            & \textbf{37.11}                  & 381                                                                   \\
\bottomrule
\end{tabular}}
\end{table*}

\subsection{Ablation Studies}
\label{ablation}
% \begin{itemize}
%     \item Test the impact of different training data construction methods.
%     \item Experiment with different training set sizes.
%     \item Provide a set of top N options.
%     \item Consider fine-tuning with FineCops-Ref training data.
In this section, we present extensive ablation studies on our proposed methods. For SFA, we analyze the impact of different detection-based difficulty-level assessment methods on performance. For CRS, we explore various training configurations, including dataset selection and the number of candidates used for region selection.

\subsubsection{\textbf{Ablation on Difficulty-level Assessment Methods in SFA}}
In SFA, we use an object-detection based strategy to assess task difficulty, which then guides model selection. Specifically, we extract instances that belong to the target category and check if there are multiple instances of that category. To evaluate this approach, we test different object detection methods, selected based on both inference time and detection capacity. The models we choose need to support open vocabulary detection due to the open-world nature of general target extraction.

We compare difficulty-level recognition accuracy and Level 2 recall on FineCops-Ref, which provides ground-truth difficulty-level labels. As shown in Table~\ref{ablation_router}, Grounding DINO~\cite{zhao2024open} achieves superior performance due to its stronger vision backbones. Accuracy indicates the precision of difficulty-level assessment, reflecting the rationality of task complexity division. Level 2 recall measures the proportion of ground-truth Level 2 tasks correctly identified. A higher Level 2 recall suggests that more complex tasks are appropriately assigned to MLLMs. While considering accuracy, it's crucial to maintain a high Level 2 recall because Specialist Models handle complex tasks poorly, and misassigning these tasks to Specialist Models can significantly degrade SFA performance. Therefore, we select MM-GDINO-L~\cite{zhao2024open} as our detection method, despite its higher computational cost, which remains within acceptable limits. 
% It is clear that there is still considerable room for improvement in our routing mechanism, and we believe the SFA strategy holds great potential for future refinement.

\begin{table}[h]
\centering
\caption{Ablation study of variation in dataset and dataset sizes in instruction tuning for CRS. $\dagger$ denotes the use of multi-choice instruction tuning.}
\label{ablation_data}
\resizebox{0.5\textwidth}{!}{
\begin{tabular}{lcccccc} 
\toprule
\textbf{Model}                                    & \textbf{Data source}         & \textbf{Data size} & \textbf{FineCops-Ref} & \textbf{Ref-Adv} & \textbf{Ref-Reasoning} \\ 
\midrule
Qwen2-VL-7B~\cite{wang2024qwen2}                                        & -                            & -                  & 73.09                 & 79.83            & 35.77                  \\ 
\midrule
\multirow{6}{*}{Qwen2-VL-7B$\dagger$} & \multirow{5}{*}{RefCOCO/+/g} & 2k                 & 74.95                 & 76.40            & 38.88                  \\
                                                  &                              & 4k                 & 76.49                 & 77.94            & 39.54                  \\
                                                  &                              & 6k                 & 77.67                 & 79.43            & 39.93                  \\
                                                  &                              & 8k                 & 78.25                 & 80.37            & 40.35                  \\
                                                  &                              & 10k                & 78.33                 & 80.56            & 40.45                  \\ 
                                                  &                              & 12k                & 78.84                 & 81.32            & 40.68                  \\ 
\cmidrule(r){2-6}
                                                  & FineCops-Ref                 & 10k                & \textbf{81.38}                 & \textbf{82.64}            & \textbf{42.05}                  \\
\bottomrule
\end{tabular}}
\end{table}

\subsubsection{\textbf{Ablation on Dataset Variation in Instruction Tuning for CRS}}
In the CRS approach, we construct datasets from RefCOCO/+/g~\cite{mao2016generation, yu2016modeling, Nagaraja2016modeling} to perform multi-choice instruction tuning. This strategy prevents test data leakage and helps the model adapt to the new output format. Table~\ref{ablation_data} shows that performance improves consistently as the dataset size increases from 2k to 12k. Additionally, we report results from training on a 10k dataset constructed from FineCops-Ref. As expected, instruction tuning on FineCops-Ref yields better performance on this dataset. More interestingly, this also leads to improved performance on the other two compositional REC datasets, Ref-Adv and Ref-Reasoning. This highlights the advantage of FineCops-Ref over traditional RefCOCO datasets in terms of enhancing the reasoning capabilities required by MLLMs for handling complex REC tasks.

%After multi-choice fine-tuning, the model surpasses the baseline and steadily improves with increasing dataset size, confirming the effectiveness of CRS in adapting the model to the new output format and yielding significant performance gains.
%Furthermore, compared to data constructed from RefCOCO, training with data from FineCops-Ref enables the model to achieve superior results across three benchmarks requiring compositional reasoning. This underscores the utility of the FineCops-Ref dataset, which includes data demanding complex reasoning. Since the primary goal of fine-tuning is instruction following, we ultimately select a small dataset size of 10k for our experimental setting. At this size, the model effectively aligns with the new output format while achieving satisfactory performance.

\begin{table}[h]
\centering
\caption{Ablation study of variation in the number of region candidates in instruction tuning for CRS. $\dagger$ denotes the use of multi-choice instruction tuning.}
\label{ablation_topk}
\resizebox{0.5\textwidth}{!}{
\begin{tabular}{lcccc} 
\toprule
\textbf{Model}                                    & \textbf{Top-k} & \textbf{FineCops-Ref} & \textbf{Ref-Adv} & \textbf{Ref-Reasoning} \\ 
\midrule
Qwen2-VL-7B~\cite{wang2024qwen2}                                        & -           & 73.09                 & 79.83            & 35.77                  \\ 
\midrule
\multirow{4}{*}{Qwen2-VL-7B$\dagger$} & Top-3          & 77.53                 & 80.16            & 39.60                  \\
                                                  & Top-5          & \textbf{78.33}                 & \textbf{80.56}            & \textbf{40.45}                  \\
                                                  & Top-8          & 77.99                 & 79.45            & 40.22                  \\
                                                  & Top-10         & 78.08                 & 80.08            & 40.24                  \\
\bottomrule
\end{tabular}}
\end{table}

\subsubsection{\textbf{Ablation on Variation in Number of Region Candidates in Instruction Tuning for CRS}}
In the multi-choice instruction tuning process of CRS, the parameter $k$ represents the number of candidate options. Specifically, the bounding boxes generated by Grounding DINO are ranked in descending order according to their confidence scores, and the top-k bounding boxes are selected as the final options. Table~\ref{ablation_topk} presents the experimental results, indicating that the method achieves optimal performance when top-5 candidates are adopted. This is due to the trade-off between recall and the introduction of distractors: while higher top-
k values increase recall, they also bring in more irrelevant candidates. Setting k=5 strikes a better balance between improving recall and minimizing distractors. Therefore, we adopt a top-k value of 5 for our experimental settings.

% \subsection{Different Specialist Models}
% \textbf{TBD.}
% \begin{itemize}
%     \item Explore the impact of varying the number of options at test time.
% \end{itemize}

% \section{In-Depth Analysis}
% \subsection{Old Analysis}
% \begin{itemize}
%     \item Present the previous analysis results.
% \end{itemize}

% \subsection{Case Study}
% \begin{itemize}
%     \item Explore the effects of adding visual prompt fine-tuning.
% \end{itemize}

\section{Conclusion}
In this work, we build upon our previously introduced FineCops-Ref dataset, which was designed for fine-grained compositional Referring Expression Comprehension (REC). FineCops-Ref features controllable difficulty levels and carefully constructed negative samples, enabling comprehensive evaluation of REC models in object detection, cross-modal reasoning, and rejection of non-existent targets—capabilities critical for real-world applications. To further advance this task, we proposed a Specialist-MLLM collaboration strategy that leverages the strengths of both object detection and cross-modal reasoning. This strategy is implemented through two approaches: Slow-Fast Adaptation (SFA), which dynamically assigns tasks to the appropriate specialist or MLLM model based on task difficulty, and Candidate Region Selection (CRS), which generates multiple region candidates using specialist models and selects the target through a multi-choice reasoning strategy. Extensive experiments on the FineCops-Ref dataset, along with two other challenging REC datasets, demonstrate the effectiveness of our proposed methods.
 
\bibliography{custom}

\begin{thebibliography}{10}
\providecommand{\url}[1]{#1}
\csname url@samestyle\endcsname
\providecommand{\newblock}{\relax}
\providecommand{\bibinfo}[2]{#2}
\providecommand{\BIBentrySTDinterwordspacing}{\spaceskip=0pt\relax}
\providecommand{\BIBentryALTinterwordstretchfactor}{4}
\providecommand{\BIBentryALTinterwordspacing}{\spaceskip=\fontdimen2\font plus
\BIBentryALTinterwordstretchfactor\fontdimen3\font minus \fontdimen4\font\relax}
\providecommand{\BIBforeignlanguage}[2]{{%
\expandafter\ifx\csname l@#1\endcsname\relax
\typeout{** WARNING: IEEEtran.bst: No hyphenation pattern has been}%
\typeout{** loaded for the language `#1'. Using the pattern for}%
\typeout{** the default language instead.}%
\else
\language=\csname l@#1\endcsname
\fi
#2}}
\providecommand{\BIBdecl}{\relax}
\BIBdecl

\bibitem{xiao2024visualgroundingsurvey}
L.~Xiao, X.~Yang, X.~Lan, Y.~Wang, and C.~Xu, ``Towards visual grounding: A survey,'' \emph{arXiv preprint arXiv:2412.20206}, 2024.

\bibitem{mao2016generation}
J.~Mao, J.~Huang, A.~Toshev, O.~Camburu, A.~L. Yuille, and K.~Murphy, ``Generation and comprehension of unambiguous object descriptions,'' in \emph{CVPR}, 2016.

\bibitem{yu2016modeling}
L.~Yu, P.~Poirson, S.~Yang, A.~C. Berg, and T.~L. Berg, ``Modeling context in referring expressions,'' in \emph{ECCV}, 2016.

\bibitem{Nagaraja2016modeling}
V.~K. Nagaraja, V.~I. Morariu, and L.~S. Davis, ``Modeling context between objects for referring expression understanding,'' in \emph{ECCV}, 2016.

\bibitem{cirik-etal-2018-visual}
V.~Cirik, L.-P. Morency, and T.~Berg-Kirkpatrick, ``Visual referring expression recognition: What do systems actually learn?'' in \emph{NAACL}, 2018.

\bibitem{akula-etal-2020-words}
A.~Akula, S.~Gella, Y.~Al-Onaizan, S.-C. Zhu, and S.~Reddy, ``Words aren{'}t enough, their order matters: On the robustness of grounding visual referring expressions,'' in \emph{ACL}, 2020.

\bibitem{liu2024finecops}
J.~Liu, X.~Yang, W.~Li, and P.~Wang, ``Finecops-ref: A new dataset and task for fine-grained compositional referring expression comprehension,'' in \emph{EMNLP}, 2024.

\bibitem{liu2019clevr}
R.~Liu, C.~Liu, Y.~Bai, and A.~L. Yuille, ``Clevr-ref+: Diagnosing visual reasoning with referring expressions,'' in \emph{CVPR}, 2019.

\bibitem{chen2020cops}
Z.~Chen, P.~Wang, L.~Ma, K.-Y.~K. Wong, and Q.~Wu, ``Cops-ref: A new dataset and task on compositional referring expression comprehension,'' in \emph{CVPR}, 2020.

\bibitem{yang2020graph}
S.~Yang, G.~Li, and Y.~Yu, ``Graph-structured referring expression reasoning in the wild,'' in \emph{CVPR}, 2020.

\bibitem{hudson2019gqa}
D.~A. Hudson and C.~D. Manning, ``Gqa: A new dataset for real-world visual reasoning and compositional question answering,'' in \emph{CVPR}, 2019.

\bibitem{Wu_2023_ICCV_GITM-MR}
Y.~Wu, Y.~Wei, H.~Wang, Y.~Liu, S.~Yang, and X.~He, ``Grounded image text matching with mismatched relation reasoning,'' in \emph{ICCV}, 2023.

\bibitem{zeng2024investigating}
Y.~Zeng, Y.~Huang, J.~Zhang, Z.~Jie, Z.~Chai, and L.~Wang, ``Investigating compositional challenges in vision-language models for visual grounding,'' in \emph{CVPR}, 2024.

\bibitem{kurita2023refego}
S.~Kurita, N.~Katsura, and E.~Onami, ``Refego: Referring expression comprehension dataset from first-person perception of ego4d,'' in \emph{ICCV}, 2023.

\bibitem{schulter2023omnilabel}
S.~Schulter, Y.~Suh, K.~M. Dafnis, Z.~Zhang, S.~Zhao, D.~Metaxas \emph{et~al.}, ``Omnilabel: A challenging benchmark for language-based object detection,'' in \emph{ICCV}, 2023.

\bibitem{shekhar-etal-2017-foil}
R.~Shekhar, S.~Pezzelle, Y.~Klimovich, A.~Herbelot, M.~Nabi, E.~Sangineto, and R.~Bernardi, ``{FOIL} it! find one mismatch between image and language caption,'' in \emph{ACL}, 2017.

\bibitem{parcalabescu-etal-2022-valse}
L.~Parcalabescu, M.~Cafagna, L.~Muradjan, A.~Frank, I.~Calixto, and A.~Gatt, ``{VALSE}: A task-independent benchmark for vision and language models centered on linguistic phenomena,'' in \emph{ACL}, 2022.

\bibitem{thrush2022winoground}
T.~Thrush, R.~Jiang, M.~Bartolo, A.~Singh, A.~Williams, D.~Kiela, and C.~Ross, ``Winoground: Probing vision and language models for visio-linguistic compositionality,'' in \emph{CVPR}, 2022.

\bibitem{ma2023crepe}
Z.~Ma, J.~Hong, M.~O. Gul, M.~Gandhi, I.~Gao, and R.~Krishna, ``Crepe: Can vision-language foundation models reason compositionally?'' in \emph{CVPR}, 2023.

\bibitem{NEURIPS2023_coco_counter}
T.~Le, V.~LAL, and P.~Howard, ``Coco-counterfactuals: Automatically constructed counterfactual examples for image-text pairs,'' in \emph{NeurIPS}, 2023.

\bibitem{Kamath2024TheHP}
A.~Kamath, C.-Y. Hsieh, K.-W. Chang, and R.~Krishna, ``The hard positive truth about vision-language compositionality,'' in \emph{ECCV}, 2024.

\bibitem{hsieh_sugarcrepe_2023}
C.-Y. Hsieh, J.~Zhang, Z.~Ma, A.~Kembhavi, and R.~Krishna, ``{SugarCrepe}: {Fixing} {Hackable} {Benchmarks} for {Vision}-{Language} {Compositionality},'' in \emph{NeurIPS}, 2023.

\bibitem{yarom_what_2023}
M.~Yarom, Y.~Bitton, S.~Changpinyo, R.~Aharoni, J.~Herzig, O.~Lang, E.~Ofek, and I.~Szpektor, ``What {You} {See} is {What} {You} {Read}? {Improving} {Text}-{Image} {Alignment} {Evaluation},'' in \emph{NeurIPS}, 2023.

\bibitem{tong2024eyes}
S.~Tong, Z.~Liu, Y.~Zhai, Y.~Ma, Y.~LeCun, and S.~Xie, ``Eyes wide shut? exploring the visual shortcomings of multimodal llms,'' in \emph{CVPR}, 2024.

\bibitem{zhang2024countercurate}
J.~Zhang, M.~Cai, T.~Xie, and Y.~J. Lee, ``Countercurate: Enhancing physical and semantic visio-linguistic compositional reasoning via counterfactual examples,'' in \emph{Findings of ACL}, 2024.

\bibitem{lin2023visualgptscore}
Z.~Lin, X.~Chen, D.~Pathak, P.~Zhang, and D.~Ramanan, ``Revisiting the role of language priors in vision-language models,'' in \emph{ICML}, 2024.

\bibitem{coco_cf}
C.~Lai, S.~Song, S.~Yan, and G.~Hu, ``Improving vision and language concepts understanding with multimodal counterfactual samples,'' in \emph{ECCV}, 2024.

\bibitem{hu2017modeling}
R.~Hu, M.~Rohrbach, J.~Andreas, T.~Darrell, and K.~Saenko, ``Modeling relationships in referential expressions with compositional modular networks,'' in \emph{CVPR}, 2017.

\bibitem{zhang2018grounding}
H.~Zhang, Y.~Niu, and S.-F. Chang, ``Grounding referring expressions in images by variational context,'' in \emph{CVPR}, 2018.

\bibitem{zhuang2018parallel}
B.~Zhuang, Q.~Wu, C.~Shen, I.~Reid, and A.~Van Den~Hengel, ``Parallel attention: A unified framework for visual object discovery through dialogs and queries,'' in \emph{CVPR}, 2018.

\bibitem{yu2018mattnet}
L.~Yu, Z.~Lin, X.~Shen, J.~Yang, X.~Lu, M.~Bansal, and T.~L. Berg, ``Mattnet: Modular attention network for referring expression comprehension,'' in \emph{CVPR}, 2018.

\bibitem{liu2019improving}
X.~Liu, Z.~Wang, J.~Shao, X.~Wang, and H.~Li, ``Improving referring expression grounding with cross-modal attention-guided erasing,'' in \emph{CVPR}, 2019.

\bibitem{yang2019dynamic}
S.~Yang, G.~Li, and Y.~Yu, ``Dynamic graph attention for referring expression comprehension,'' in \emph{ICCV}, 2019.

\bibitem{hong2019learning}
R.~Hong, D.~Liu, X.~Mo, X.~He, and H.~Zhang, ``Learning to compose and reason with language tree structures for visual grounding,'' \emph{TPAMI}, 2019.

\bibitem{liu2019learning}
D.~Liu, H.~Zhang, F.~Wu, and Z.-J. Zha, ``Learning to assemble neural module tree networks for visual grounding,'' in \emph{ICCV}, 2019.

\bibitem{Wang2018NeighbourhoodWR}
P.~Wang, Q.~Wu, J.~Cao, C.~Shen, L.~Gao, and A.~van~den Hengel, ``Neighbourhood watch: Referring expression comprehension via language-guided graph attention networks,'' in \emph{CVPR}, 2018.

\bibitem{zhou2021real}
Y.~Zhou, R.~Ji, G.~Luo, X.~Sun, J.~Su, X.~Ding, C.-W. Lin, and Q.~Tian, ``A real-time global inference network for one-stage referring expression comprehension,'' \emph{TNNLS}, 2021.

\bibitem{deng2021transvg}
J.~Deng, Z.~Yang, T.~Chen, W.~Zhou, and H.~Li, ``Transvg: End-to-end visual grounding with transformers,'' in \emph{ICCV}, 2021.

\bibitem{zhu2022seqtr}
C.~Zhu, Y.~Zhou, Y.~Shen, G.~Luo, X.~Pan, M.~Lin, C.~Chen, L.~Cao, X.~Sun, and R.~Ji, ``Seqtr: A simple yet universal network for visual grounding,'' in \emph{ECCV}, 2022.

\bibitem{TransVG++2023}
J.~Deng, Z.~Yang, D.~Liu, T.~Chen, W.~Zhou, Y.~Zhang, H.~Li, and W.~Ouyang, ``Transvg++: End-to-end visual grounding with language conditioned vision transformer,'' \emph{TPAMI}, 2023.

\bibitem{kamath2021mdetr}
A.~Kamath, M.~Singh, Y.~LeCun, G.~Synnaeve, I.~Misra, and N.~Carion, ``Mdetr-modulated detection for end-to-end multi-modal understanding,'' in \emph{ICCV}, 2021.

\bibitem{yan2023universal}
B.~Yan, Y.~Jiang, J.~Wu, D.~Wang, P.~Luo, Z.~Yuan, and H.~Lu, ``Universal instance perception as object discovery and retrieval,'' in \emph{CVPR}, 2023.

\bibitem{liu2023grounding}
S.~Liu, Z.~Zeng, T.~Ren, F.~Li, H.~Zhang, J.~Yang, Q.~Jiang, C.~Li, J.~Yang, H.~Su \emph{et~al.}, ``Grounding dino: Marrying dino with grounded pre-training for open-set object detection,'' in \emph{ECCV}, 2024.

\bibitem{dMDETR23}
F.~Shi, R.~Gao, W.~Huang, and L.~Wang, ``Dynamic mdetr: A dynamic multimodal transformer decoder for visual grounding,'' \emph{TPAMI}, 2024.

\bibitem{carion2020end}
N.~Carion, F.~Massa, G.~Synnaeve, N.~Usunier, A.~Kirillov, and S.~Zagoruyko, ``End-to-end object detection with transformers,'' in \emph{ECCV}, 2020.

\bibitem{zhu2020deformable}
X.~Zhu, W.~Su, L.~Lu, B.~Li, X.~Wang, and J.~Dai, ``Deformable detr: Deformable transformers for end-to-end object detection,'' in \emph{ICLR}, 2021.

\bibitem{zhang2022dino}
H.~Zhang, F.~Li, S.~Liu, L.~Zhang, H.~Su, J.~Zhu, L.~Ni, and H.-Y. Shum, ``{DINO}: {DETR} with improved denoising anchor boxes for end-to-end object detection,'' in \emph{ICLR}, 2023.

\bibitem{Zhao_2024_CVPR}
S.~Zhao, L.~Zhao, V.~K.~B. G, Y.~Suh, D.~N. Metaxas, M.~Chandraker, and S.~Schulter, ``Generating enhanced negatives for training language-based object detectors,'' in \emph{CVPR}, 2024.

\bibitem{xiao2024oneref}
L.~Xiao, X.~Yang, F.~Peng, Y.~Wang, and C.~Xu, ``Oneref: Unified one-tower expression grounding and segmentation with mask referring modeling,'' in \emph{NeurIPS}, 2024.

\bibitem{dai2024simvg}
M.~Dai, L.~Yang, Y.~Xu, Z.~Feng, and W.~Yang, ``Simvg: A simple framework for visual grounding with decoupled multi-modal fusion,'' in \emph{NeurIPS}, 2024.

\bibitem{chen2023shikra}
K.~Chen, Z.~Zhang, W.~Zeng, R.~Zhang, F.~Zhu, and R.~Zhao, ``Shikra: Unleashing multimodal llm's referential dialogue magic,'' \emph{arXiv preprint arXiv:2306.15195}, 2023.

\bibitem{you2024ferret}
H.~You, H.~Zhang, Z.~Gan, X.~Du, B.~Zhang, Z.~Wang, L.~Cao, S.-F. Chang, and Y.~Yang, ``Ferret: Refer and ground anything anywhere at any granularity,'' in \emph{ICLR}, 2024.

\bibitem{li-etal-2024-groundinggpt}
Z.~Li, Q.~Xu, D.~Zhang, H.~Song, Y.~Cai, Q.~Qi, R.~Zhou, J.~Pan, Z.~Li, V.~Tu, Z.~Huang, and T.~Wang, ``{G}rounding{GPT}: Language enhanced multi-modal grounding model,'' in \emph{ACL}, 2024.

\bibitem{wei2023lenna}
F.~Wei, X.~Zhang, A.~Zhang, B.~Zhang, and X.~Chu, ``Lenna: Language enhanced reasoning detection assistant,'' \emph{arXiv preprint arXiv:2312.02433}, 2023.

\bibitem{wang2023cogvlm}
W.~Wang, Q.~Lv, W.~Yu, W.~Hong, J.~Qi, Y.~Wang, J.~Ji, Z.~Yang, L.~Zhao, X.~Song \emph{et~al.}, ``Cogvlm: Visual expert for pretrained language models,'' in \emph{NeurIPS}, 2024.

\bibitem{qi2024cogcom}
J.~Qi, M.~Ding, W.~Wang, Y.~Bai, Q.~Lv, W.~Hong, B.~Xu, L.~Hou, J.~Li, Y.~Dong \emph{et~al.}, ``Cogcom: Train large vision-language models diving into details through chain of manipulations,'' \emph{arXiv preprint arXiv:2402.04236}, 2024.

\bibitem{chen2024far}
Z.~Chen, W.~Wang, H.~Tian, S.~Ye, Z.~Gao, E.~Cui, W.~Tong, K.~Hu, J.~Luo, Z.~Ma \emph{et~al.}, ``How far are we to gpt-4v? closing the gap to commercial multimodal models with open-source suites,'' \emph{Science China Information Sciences}, 2024.

\bibitem{yang2023set}
J.~Yang, H.~Zhang, F.~Li, X.~Zou, C.~Li, and J.~Gao, ``Set-of-mark prompting unleashes extraordinary visual grounding in gpt-4v,'' \emph{arXiv preprint arXiv:2310.11441}, 2023.

\bibitem{Rasheed_2024_CVPR}
H.~Rasheed, M.~Maaz, S.~Shaji, A.~Shaker, S.~Khan, H.~Cholakkal, R.~M. Anwer, E.~Xing, M.-H. Yang, and F.~S. Khan, ``Glamm: Pixel grounding large multimodal model,'' in \emph{CVPR}, 2024.

\bibitem{zhang2023llava-g}
H.~Zhang, H.~Li, F.~Li, T.~Ren, X.~Zou, S.~Liu, S.~Huang, J.~Gao, Leizhang, C.~Li \emph{et~al.}, ``Llava-grounding: Grounded visual chat with large multimodal models,'' in \emph{ECCV}, 2024.

\bibitem{ma2024groma}
C.~Ma, Y.~Jiang, J.~Wu, Z.~Yuan, and X.~Qi, ``Groma: Localized visual tokenization for grounding multimodal large language models,'' in \emph{ECCV}, 2024.

\bibitem{shao2024visualcot}
H.~Shao, S.~Qian, H.~Xiao, G.~Song, Z.~ZONG, L.~Wang, Y.~Liu, and H.~Li, ``Visual cot: Advancing multi-modal language models with a comprehensive dataset and benchmark for chain-of-thought reasoning,'' in \emph{NeurIPS}, 2024.

\bibitem{li2024covlm}
J.~Li, D.~Chen, Y.~Hong, Z.~Chen, P.~Chen, Y.~Shen, and C.~Gan, ``Co{VLM}: Composing visual entities and relationships in large language models via communicative decoding,'' in \emph{ICLR}, 2024.

\bibitem{wang2024qwen2}
P.~Wang, S.~Bai, S.~Tan, S.~Wang, Z.~Fan, J.~Bai, K.~Chen, X.~Liu, J.~Wang, W.~Ge \emph{et~al.}, ``Qwen2-vl: Enhancing vision-language model's perception of the world at any resolution,'' \emph{arXiv preprint arXiv:2409.12191}, 2024.

\bibitem{chen2024expanding}
Z.~Chen, W.~Wang, Y.~Cao, Y.~Liu, Z.~Gao, E.~Cui, J.~Zhu, S.~Ye, H.~Tian, Z.~Liu \emph{et~al.}, ``Expanding performance boundaries of open-source multimodal models with model, data, and test-time scaling,'' \emph{arXiv preprint arXiv:2412.05271}, 2024.

\bibitem{zhao2024open}
X.~Zhao, Y.~Chen, S.~Xu, X.~Li, X.~Wang, Y.~Li, and H.~Huang, ``An open and comprehensive pipeline for unified object grounding and detection,'' \emph{arXiv preprint arXiv:2401.02361}, 2024.

\bibitem{achiam2023gpt}
J.~Achiam, S.~Adler, S.~Agarwal, L.~Ahmad, I.~Akkaya, F.~L. Aleman, D.~Almeida, J.~Altenschmidt, S.~Altman, S.~Anadkat \emph{et~al.}, ``Gpt-4 technical report,'' \emph{arXiv preprint arXiv:2303.08774}, 2023.

\bibitem{mitchell2023detectgpt}
E.~Mitchell, Y.~Lee, A.~Khazatsky, C.~D. Manning, and C.~Finn, ``Detectgpt: Zero-shot machine-generated text detection using probability curvature,'' in \emph{ICML}, 2023.

\bibitem{wang2024ov}
H.~Wang, P.~Ren, Z.~Jie, X.~Dong, C.~Feng, Y.~Qian, L.~Ma, D.~Jiang, Y.~Wang, X.~Lan \emph{et~al.}, ``Ov-dino: Unified open-vocabulary detection with language-aware selective fusion,'' \emph{arXiv preprint arXiv:2407.07844}, 2024.

\bibitem{jeong2024proxydet}
J.~Jeong, G.~Park, J.~Yoo, H.~Jung, and H.~Kim, ``Proxydet: Synthesizing proxy novel classes via classwise mixup for open-vocabulary object detection,'' in \emph{AAAI}, 2024.

\bibitem{cheng2024yolo}
T.~Cheng, L.~Song, Y.~Ge, W.~Liu, X.~Wang, and Y.~Shan, ``Yolo-world: Real-time open-vocabulary object detection,'' in \emph{CVPR}, 2024.

\end{thebibliography}
\bibliographystyle{IEEEtran}

\end{document}